\documentclass{article}

\PassOptionsToPackage{numbers,sort&compress}{natbib}
\usepackage[preprint]{neurips_2026}

\usepackage[utf8]{inputenc} % allow utf-8 input
\usepackage[T1]{fontenc}    % use 8-bit T1 fonts
\usepackage{graphicx}       % graphics and resizebox
\usepackage{float}          % fixed-position figures
\usepackage{svg}            % svg figures
\usepackage{hyperref}       % hyperlinks
\usepackage{url}            % simple URL typesetting
\usepackage{booktabs}       % professional-quality tables
\usepackage{multirow}       % multi-row cells in tables
\usepackage{amsmath}        % math environments
\usepackage{amsfonts}       % blackboard math symbols
\usepackage{nicefrac}       % compact symbols for 1/2, etc.
\usepackage{microtype}      % microtypography
\usepackage{xcolor}         % colors

\title{VICR: Visual In-Context Restoration for Real-World Image Super-Resolution}

\author{
{\bfseries Qichang Zhang$^{1,2}$ \quad
Hailong Wang$^{2}$ \quad
Baiang Li$^{3}$ \quad
Linhao Wang$^{4}$}\\
{\bfseries Rong Fu$^{1}$ \quad
Erkang Cheng$^{2}$ \quad
Simon James Fong$^{1}$}\\
{\normalfont $^{1}$Faculty of Science and Technology, University of Macau}\\
{\normalfont $^{2}$Nullmax \quad
$^{3}$Hefei University of Technology \quad
$^{4}$Shandong Normal University}\\
{\normalfont \texttt{mc55084@um.edu.mo}}
}

\begin{document}

\maketitle

\begin{abstract}
  Real-world image super-resolution (Real-ISR) requires balancing structural fidelity to degraded observations with realistic detail synthesis. However, existing generative Real-ISR methods often rely on entangled conditioning mechanisms, leading to structural drift or semantically inconsistent details. To address this issue, we propose Visual In-Context Restoration (VICR), a Diffusion Transformer (DiT)-based framework that formulates Real-ISR as image completion. Specifically, we introduce a decoupled visual prior injection mechanism that derives local and global cues from the low-quality (LQ) image: local cues help recover image structures and support high-frequency detail synthesis, while global cues guide overall generation and promote semantic consistency. For ambiguous regions under severe degradation, VICR employs an inference-time agent to refine semantic prompts using visual evidence from the LQ input while keeping model parameters fixed. Experiments show that VICR achieves state-of-the-art performance across multiple Real-ISR benchmarks with only 127M trainable parameters.

\end{abstract}

\section{Introduction}

Real-world image super-resolution (Real-ISR) aims to recover high-resolution (HR) images from low-quality (LQ) inputs with unknown and compound degradations~\citep{wang2021realesrgan,wang2023stablesr,wu2024seesr}. With the emergence of large-scale generative models, super-resolution (SR) has shifted from deterministic reconstruction toward conditional generation under severe information uncertainty. In this new regime, the goal is no longer limited to recovering observable structures, but also to synthesizing plausible details that are not explicitly present in the degraded input. This introduces a fundamental challenge: how to coordinate incomplete visual evidence with strong generative priors, so that plausible details can be synthesized without sacrificing structural fidelity.

\begin{figure}[t]
  \centering
  \includegraphics[width=\textwidth]{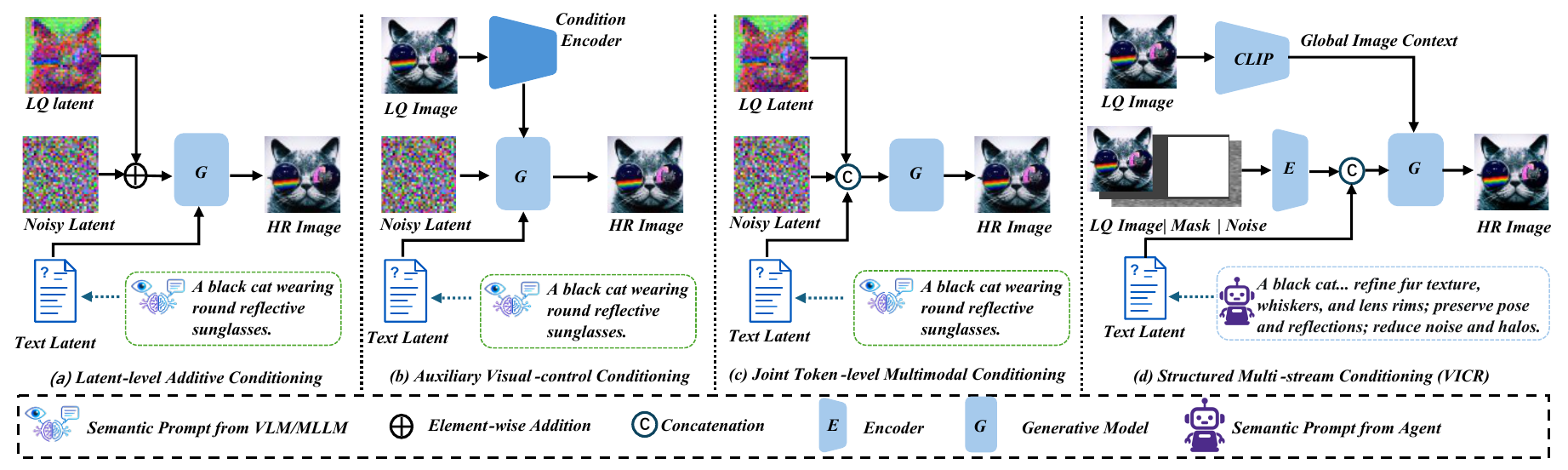}
  \caption{Schematic comparison of conditioning paradigms in generative Real-ISR\@. Prior methods typically inject low-quality observations and semantic prompts through additive latent conditioning, auxiliary visual-control branches, or joint token-level multimodal conditioning. VICR instead organizes local visual evidence, global image context, and agent-refined semantic prompts as distinct condition streams with complementary functions for the SR objective.}
  \label{fig:related_paradigms}
\end{figure}

Recent generative Real-ISR methods typically guide diffusion models with low-quality images, extracted visual features, or semantic prompts~\citep{wang2023stablesr,lin2024diffbir,yang2024pasd,wu2024seesr,wu2024osediff,duan2025dit4sr}. These conditions provide useful constraints, but they are often merged into a single control pathway, such as latent-level additive conditioning~\citep{wu2024osediff,chen2025faithdiff,kong2025dpir}, auxiliary control branches~\citep{wang2023stablesr,lin2024diffbir,yang2024pasd,wu2024seesr,ai2024dreamclear}, or joint token-level conditioning~\citep{duan2025dit4sr}, as illustrated in the first three panels of Figure~\ref{fig:related_paradigms}. Consequently, heterogeneous cues with distinct roles in the SR process are mixed together, leaving the model to coordinate them implicitly during generation. Under complex degradations, this implicit coordination can be unreliable, leading to suboptimal trade-offs, such as structural drift, hallucinated semantics, or unstable local details. This motivates a conditioning design that explicitly separates heterogeneous cues according to their roles in the SR process, rather than simply aggregating them as generic control signals. We provide a more detailed discussion of prior conditioning paradigms, DiT-based condition organization, and semantic prompting under severe degradations in Appendix~\ref{app:condition_modeling}.

To address this problem, we propose Visual In-Context Restoration (VICR), a Diffusion Transformer (DiT)-based Real-ISR framework that reformulates Real-ISR as visual in-context completion. Our key insight is that effective Real-ISR requires explicit decomposition and role assignment of heterogeneous conditions, rather than entangled condition injection. Based on this principle, VICR organizes three complementary condition streams, as illustrated in the fourth panel of Figure~\ref{fig:related_paradigms}: (1) local visual evidence, which serves as a spatially aligned reference for recovering observable structures and fine details; (2) global image context, which provides scene-level guidance and stabilizes low-frequency structure; and (3) semantic prompts, which supply instance-specific high-level information to resolve ambiguity in severely degraded regions. To realize this design, we introduce a Decoupled Prior Injection mechanism that separates local visual evidence from global image context within a unified DiT-based generative framework. In this mechanism, local visual evidence is provided through a diptych, a side-by-side image pair where the upsampled low-quality image on the left serves as visible context and the right side defines the target SR region to be completed, while a lightweight Vision-Context Bridge adapts CLIP visual features to provide global image context. In addition, we propose an inference-time agent-driven prompting strategy, which adaptively refines semantic prompts based on both the degraded input and intermediate SR results. Unlike prior approaches that rely on static prompts~\citep{chen2023promptsr,yang2024pasd,wu2024seesr,jiang2024dalpsr,chen2025srsr}, our method treats semantic prompts as a dynamic and instance-specific process, enabling more accurate and controllable detail synthesis without updating model parameters.

By explicitly structuring condition interactions, VICR enables more stable coordination between local visual evidence, global image context, and semantic prompts. Extensive experiments demonstrate that our approach achieves state-of-the-art performance on multiple Real-ISR benchmarks, while requiring significantly fewer trainable parameters. These results highlight the importance of structured condition modeling as a key principle for advancing generative Real-ISR.

The main contributions of this paper are summarized as follows:
{\setlength{\parskip}{0pt}
\setlength{\topsep}{0pt}
\setlength{\partopsep}{0pt}
\setlength{\itemsep}{0pt}
\setlength{\parsep}{0pt}
\begin{itemize}
  \item[\textcolor{black}{$\bullet$}] We propose VICR, a DiT-based Real-ISR framework that formulates Real-ISR as visual in-context completion and explicitly organizes heterogeneous conditions according to their roles in the SR process.
  \item[\textcolor{black}{$\bullet$}] We introduce Decoupled Prior Injection, which separates local visual evidence and global image context within a unified generative framework.
  \item[\textcolor{black}{$\bullet$}] We propose an inference-time agent-driven prompting strategy that adaptively refines semantic prompts based on the degraded input and intermediate SR results without updating model parameters.
  \item[\textcolor{black}{$\bullet$}] Extensive experiments on multiple Real-ISR benchmarks show that VICR achieves state-of-the-art performance with only 127M trainable adaptation parameters.
\end{itemize}
}

\section{Related work}

\noindent\textbf{Real-world Image Super-Resolution.}
Real-ISR aims to recover HR images from LQ inputs with complex and unknown degradations.
Reconstruction-oriented methods~\citep{dong2016srcnn,lim2017edsr,zhang2018rcan,liang2021swinir,conde2022swin2sr} mainly optimize fidelity under simplified degradation assumptions, while realistic-degradation training and GAN-based objectives~\citep{cai2019realsr,zhang2021bsrgan,wang2021realesrgan,wei2021dasr,ledig2017srgan,sajjadi2017enhancenet,wang2018sftgan,wang2018esrgan,zhang2019ranksrgan,fritsche2019frequency,maeda2020unpaired} improve robustness and perceptual realism but may introduce artifacts under severe degradation.
Recent diffusion-based Real-ISR methods exploit pretrained diffusion or text-to-image priors through latent restoration~\citep{wang2023stablesr}, two-stage restoration and regeneration~\citep{lin2024diffbir}, pixel-aware or faithful prior injection~\citep{yang2024pasd,chen2025faithdiff}, restoration-specific scaling, one-step adaptation, or decoder adaptation~\citep{yu2024supir,wu2024osediff,yi2025tvt}, and more recently DiT-based restoration frameworks~\citep{ai2024dreamclear,kong2025dpir,duan2025dit4sr}.
Most existing diffusion-based methods inject LQ observations through latent-level additive conditioning or auxiliary visual-control pathways, and introduce semantic cues through text or multimodal attention.
However, these heterogeneous cues are typically fused by generic conditioning mechanisms, leaving the roles of local visual evidence, global image context, and semantic prompts weakly separated.

\noindent\textbf{Semantic Prompting for Image Super-Resolution.}
Severe degradations destroy local visual evidence, so a parallel line of work injects language to disambiguate scene and object semantics. PASD~\citep{yang2024pasd} couples pixel-aware attention with classification, detection, and captioning signals. SeeSR~\citep{wu2024seesr} trains a degradation-aware extractor producing hard tags and soft prompts, while SUPIR~\citep{yu2024supir} uses MLLM-generated and user prompts as a control interface. Recent works further spatialize or decouple textual priors: DTPSR~\citep{jiang2026dtpsr} separates textual priors by spatial hierarchy and frequency, while SRSR~\citep{chen2025srsr} grounds cross-attention with masks. Unlike fixed or one-shot prompts, our prompts are iteratively refined by an agent that inspects intermediate SR results. Nevertheless, language remains descriptive rather than evidential: it can clarify semantics but cannot specify instance-specific edges, contours, spatial layouts, or textures. VICR therefore assigns such visual evidence to a separate condition rather than overloading semantic prompts.

\noindent\textbf{Visual In-Context Generation and Restoration.}
Recent DiT/FLUX in-context systems use images directly as in-context demonstrations, without routing every visual condition through an adapter. FLUX.1 Fill~\citep{flux2024fill} provides a rectified-flow inpainting interface, and FLUX.1 Kontext~\citep{flux2025kontext} unifies image and text inputs via sequence concatenation. In-Context LoRA~\citep{huang2024iclora} shows that a single shared canvas already elicits strong cross-image contextual modeling. Diptych Prompting~\citep{shin2024diptych} formalizes subject-driven generation as inpainting an incomplete diptych with a visible reference panel and a masked target panel. ICEdit~\citep{zhang2025icedit} uses in-context generation for instruction-based image editing, and Insert Anything~\citep{song2025insertanything}, In-Context Brush~\citep{xu2025incontextbrush}, and IC-Custom~\citep{li2025iccustom} extend it to reference-based insertion and customization. VisualCloze~\citep{li2025visualcloze} generalizes the pattern to a unified visual in-context completion framework. None of these methods address Real-ISR\@. VICR brings the construction to Real-ISR\@. The upsampled LQ image becomes the visible panel and the SR output is the masked panel. The diptych carries local visual evidence with explicit spatial correspondence and removes the need for a separate local encoder.

\begin{figure}[t]
  \centering
  \includegraphics[width=\textwidth]{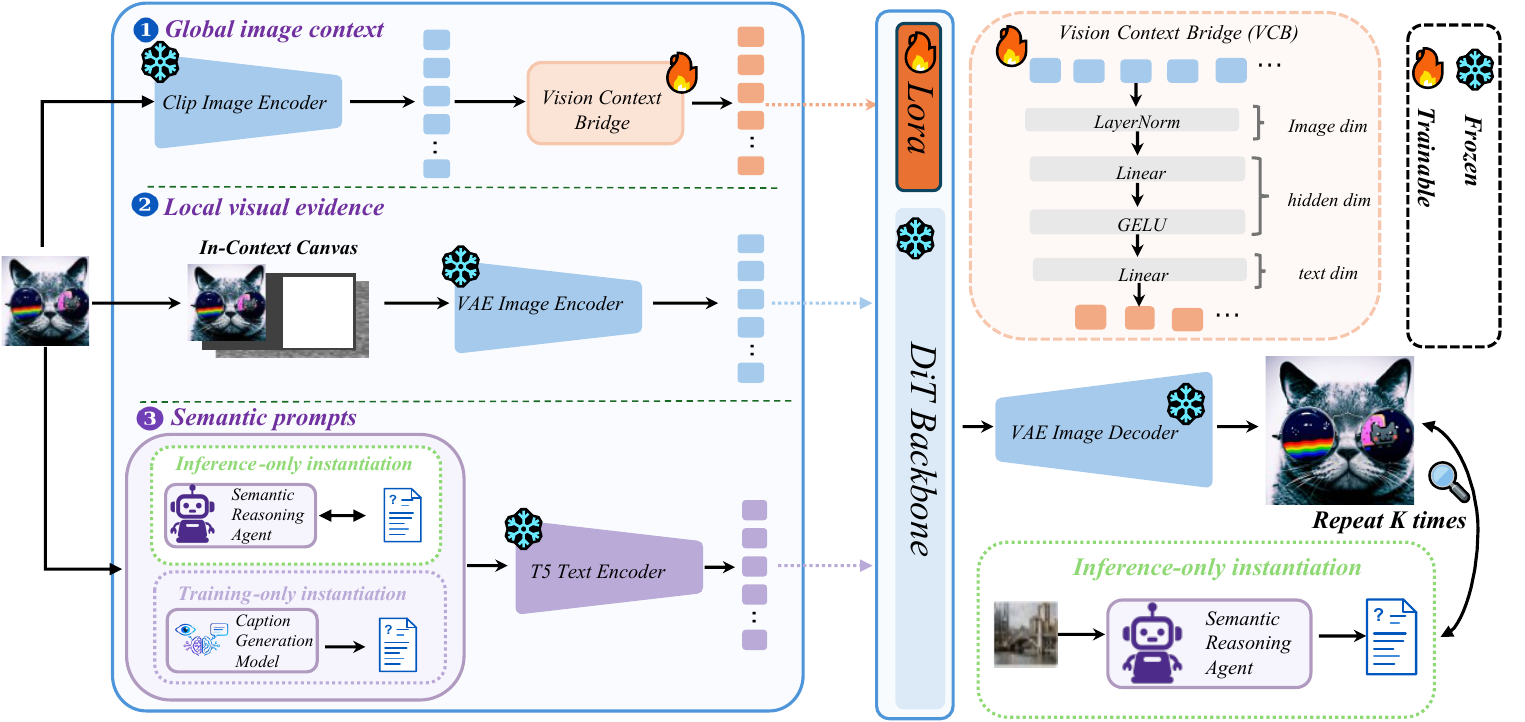}
  \caption{Overview of VICR\@. Real-ISR is cast as visual in-context completion. Local visual evidence comes from the diptych, global image context from the Vision-Context Bridge over CLIP, and semantic prompts from LLaVA captions during training and agent refinement during inference.}
  \label{fig:vicr_pipeline}
\end{figure}

\section{Method}

\subsection{Overall framework}

Many diffusion-based Real-ISR methods couple low-quality observations, visual representations, and semantic cues within a shared conditioning interface. Under this coupled design, feature-map adapters can weaken spatial correspondence between the LQ input and SR output, while injecting visual and textual conditions through the same pathway makes structural fidelity and semantic stability compete. VICR removes this entanglement by assigning each condition a separate role.

VICR is a DiT-based Real-ISR framework. Given a low-quality input \(x_{lq}\), we denote the SR process as
\begin{equation}
\hat{y}=\mathcal{G}_{\theta}(c_v^{\mathrm{local}}, c_v^{\mathrm{global}}, c_t),
\end{equation}
where \(\mathcal{G}_{\theta}\) is a Flux-Fill-based generative backbone~\citep{flux2024fill}, \(c_v^{\mathrm{local}}\) is the diptych-derived local visual condition, \(c_v^{\mathrm{global}}\) is a global image context vector, and \(c_t\) is the textual condition.

\begin{figure}[t]
  \centering
  \includegraphics[width=\textwidth]{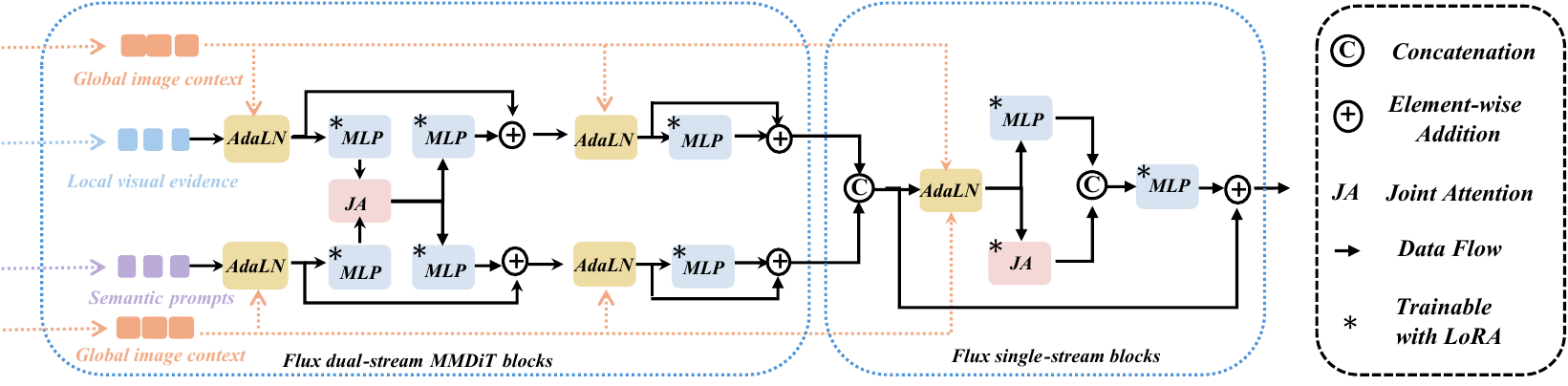}
  \caption{Detailed condition injection in VICR\@. Global image context, local visual evidence, and semantic prompts are routed through the Flux dual-stream and single-stream MMDiT blocks, while lightweight LoRA adaptation updates the trainable modulation and projection layers.}
  \label{fig:vicr_pipeline_detail}
\end{figure}

The local stream preserves structural details and per-pixel correspondence to the LQ input, the global stream stabilizes scene-level layout and low-frequency content, and the semantic prompt compensates for high-level ambiguity when local evidence is severely degraded. Figure~\ref{fig:vicr_pipeline} gives an overview of the VICR pipeline, while Figure~\ref{fig:vicr_pipeline_detail} details how the local, global, and semantic condition streams are injected into the Flux MMDiT blocks. The construction of each stream is described in the following subsections.

\subsection{Local visual evidence via diptych completion}

Encoding the LQ image with a separate encoder or control branch compresses local structure into abstract tokens, weakens spatial correspondence, and makes the model rely on the adapter to relay every edge and contour. We avoid an extra local encoder. Real-ISR is cast as inpainting-based in-context completion. Let \(U_{\mathrm{bic}}\) denote bicubic upsampling, and let \(\operatorname{cat}_w(\cdot,\cdot)\) denote concatenation along the width dimension. The bicubic-upsampled LQ image \(x^{up}=U_{\mathrm{bic}}(x_{lq})\) occupies the left panel as visible context, and the right panel is masked as the SR target. During training, we build width-concatenated diptych tensors and zero the right half of the base image:
\begin{equation}
d_{\mathrm{tgt}} = \operatorname{cat}_w(x^{up}, y), \quad
d_{\mathrm{base}} = \operatorname{cat}_w(x^{up}, x^{up}), \quad
m = \operatorname{cat}_w(\mathbf{0}, \mathbf{1}),
\end{equation}
where \(y\) is the HR target during training, \(d_{\mathrm{tgt}}\) and \(d_{\mathrm{base}}\) are the target and base diptych images, \(m\) is the right-panel fill mask, and the masked diptych is \(x_{\mathrm{mi}} = d_{\mathrm{base}}\odot(1-m)\). The transformer streams are then obtained by VAE encoding and patch packing following the Flux Fill interface~\citep{flux2024fill}. We denote the resulting packed local visual condition as \(c_v^{\mathrm{local}}\).
Local structures, contours, and textures stay visible to the backbone as in-context evidence rather than being compressed into control tokens, which preserves spatial correspondence between input and output. Casting Real-ISR as image completion also follows the inpainting and visual in-context generation settings used by recent DiT systems~\citep{flux2025kontext,zhang2025icedit,shin2024diptych,li2025visualcloze}. The local-only configuration is already a strong base setting (Table~\ref{tab:ablation_vicr_reallr200}); the global and semantic streams are added on top.

\subsection{Global image context via the Vision-Context Bridge}

The diptych preserves observable structure but does not give the backbone a compact image-level
descriptor of color, illumination, and scene layout. Heavy degradations introduce globally inconsistent
cues, such as color cast, blur direction, and scene-level lighting, that the diptych does not
surface as a coherent prior. We therefore add a global stream whose only role is scene-level low-frequency
stabilization, not local texture synthesis.

To bridge the CLIP visual feature space~\citep{radford2021clip} and the conditioning space of
\(\mathcal{G}_{\theta}\), the lightweight Vision-Context Bridge (VCB) adapts the global representation
of \(x_{lq}\). We first extract an image-level visual descriptor using a frozen CLIP image encoder:
\begin{equation}
g_v = \operatorname{Pool}\!\left(\Phi_{\mathrm{CLIP}}(x_{lq})\right),
\end{equation}
where \(\Phi_{\mathrm{CLIP}}\) denotes the CLIP image encoder and
\(\operatorname{Pool}(\cdot)\) extracts the global image representation. The VCB then maps this
descriptor into the conditioning space:
\begin{equation}
c_v^{\mathrm{global}}
=
\operatorname{VCB}(g_v)
=
s \cdot W_2\,\sigma\!\left(W_1\,\operatorname{LN}(g_v)\right),
\end{equation}
where \(\operatorname{LN}(\cdot)\) is LayerNorm, \(\sigma(\cdot)\) is GELU, \(W_1\) and \(W_2\)
are linear projections, and \(s\) is a learnable scale. We inject \(c_v^{\mathrm{global}}\) as a pooled
image-conditioning vector alongside the text-conditioning pathway, not as spatially-aligned tokens,
which keeps it from competing with the diptych for local geometry. Adding the global stream on top
of the diptych delivers the largest single gain (LIQE \(3.96\!\rightarrow\!4.23\), MUSIQ
\(68.10\!\rightarrow\!69.82\) on RealLR200), consistent with its role as a scene-level stabilizer that
the local branch cannot easily provide.

\subsection{Inference-time agent-driven prompt refinement}

When the LQ input is severely degraded, local visual evidence may be insufficient to determine
object categories, scene attributes, or fine semantic details. VICR therefore uses semantic prompts
to provide high-level guidance. The prompts do not replace local visual evidence; instead, they
describe the scene and objects to help resolve ambiguous regions.

At inference time, a LLaVA-generated caption initializes the semantic prompt \(p^{(0)}\),
and the corresponding SR result \(\hat{y}^{(0)}\) serves as the static-prompt baseline.
A multimodal agent then refines only the semantic prompt across generation rounds.

For the \(k\)-th refinement round, the agent compares the previous SR result with the LQ input,
reviews ambiguous, over-smoothed, or structurally inconsistent regions, and revises only the semantic prompt:
\begin{equation}
p^{(k)}
=
\mathcal{A}
\!\left(x_{lq}, \hat{y}^{(k-1)}, p^{(k-1)}\right),
\qquad k=1,\ldots,K.
\end{equation}
Given \(p^{(k)}\), we encode it into \(c_t^{(k)}=\mathcal{E}_{t}(p^{(k)})\), and generate
the corresponding SR candidate by
\begin{equation}
\hat{y}^{(k)}
=
\mathcal{G}_{\theta}
\!\left(c_v^{\mathrm{local}}, c_v^{\mathrm{global}}, c_t^{(k)};\epsilon\right),
\qquad k=1,\ldots,K.
\end{equation}
Here \(K\) denotes the number of agent-refined generation rounds, \(\mathcal{A}\) denotes the prompt
refinement agent, and \(\mathcal{E}_{t}\) is the text encoder.

After obtaining \(K\) agent-refined candidates \(\{\hat{y}^{(k)}\}_{k=1}^{K}\), we perform
per-image Best-of-\(K\) selection as a post-processing step. For each image, we rank its \(K\)
candidates separately by LIQE, MUSIQ, MANIQA, and CLIPIQA, with rank 1 assigned to the best
candidate under each metric. We select the candidate with the smallest sum of the four ranks.
Ties are broken by a four-metric no-reference score obtained by min--max normalizing the four
metrics within the same candidate set and averaging them; if a tie still remains, the earlier
refinement round is selected.
This no-reference selection is used because HR targets are unavailable in the target Real-ISR test setting, so the \(K\) candidates cannot be compared by full-reference metrics. We use LIQE, MUSIQ, MANIQA, and CLIPIQA together, and select the candidate with the best overall rank, to avoid relying on a single no-reference metric.

\subsection{Training and inference strategy}

We keep the visual condition construction fixed and use natural-language captions as the default textual supervision. During training, captions are produced by LLaVA~\citep{liu2023llava} from the LQ input. We additionally support a discrete-tag variant from a degradation-aware tag extractor (DAPE-style); we use it in the ablation to analyze textual supervision. All main results use LLaVA captions. Training follows the flow-matching objective, where \(v_{\theta}\) denotes the velocity predictor of \(\mathcal{G}_{\theta}\):
\begin{equation}
\mathcal{L}_{\mathrm{FM}}
=
\mathbb{E}_{x,\epsilon,t}
\left[
\left\|
m_z \odot
\left(
v_{\theta}(z_t, t, c_v^{\mathrm{local}}, c_v^{\mathrm{global}}, c_t)
-
(\epsilon - x)
\right)
\right\|_2^2
\right],
\qquad
z_t = (1-t)x + t\epsilon,
\end{equation}
where \(x\) is the target diptych latent, \(m_z\) denotes \(m\) resized and packed to match the latent layout of \(z_t\), with ones on the SR target panel and zeros on the visible context panel, \(\epsilon\) is Gaussian noise, \(t\) is the sampled time step, and \(z_t\) is the interpolated latent. Only the masked right panel is supervised; gradients on the visible left panel are masked out, so the model learns to complete the SR output rather than reproduce its own input.

At inference, the textual condition is initialized by the LLaVA caption and then refined by GPT-5.4. Each refinement call uses a fixed instruction template with the LQ image, the previous-round raw SR result, and the previous prompt as inputs, and returns a full revised prompt rather than a local patch; the full instruction template and example prompt edits are provided in Appendix~\ref{app:agent_prompt_ablation}. The inference stage keeps \(\mathcal{G}_{\theta}\), \(c_v^{\mathrm{local}}\), \(c_v^{\mathrm{global}}\), and the sampling noise \(\epsilon\) fixed, uses no manually selected crops, human edits, or external web search, and constrains the agent to mark uncertain details conservatively instead of adding unsupported identities, readable text, or object attributes. Best-of-\(K\) selection is also test-time only and does not update model parameters. Unless otherwise specified, the main VICR results use \(K=10\) agent-refined candidates with Best-of-\(K\) selection.

\section{Experiments}

\subsection{Experimental settings}

\noindent\textbf{Training datasets.}
Following the training protocol of SeeSR~\citep{wu2024seesr}, we train VICR on LSDIR~\citep{li2023lsdir} and the first 10K face images from FFHQ~\citep{karras2019stylegan}. Low-resolution/high-resolution (LR-HR) training pairs are synthesized with the Real-ESRGAN degradation pipeline~\citep{wang2021realesrgan} under the \(4\times\) SR setting.

\noindent\textbf{Testing datasets.}
We evaluate VICR on three standard \(4\times\) SR benchmarks preprocessed by StableSR~\citep{wang2023stablesr}: DIV2K~\citep{agustsson2017div2k}, RealSR~\citep{cai2019realsr}, and DRealSR~\citep{wei2020drealsr}, using the LR/HR pairs provided by StableSR\@. Among them, DIV2K is a synthetic benchmark, while RealSR and DRealSR are real-world SR benchmarks\@. To further assess generalization on real images without reference targets, we additionally report no-reference results on RealLR200~\citep{wu2024seesr} and RealLQ250~\citep{wang2023stablesr}.

\noindent\textbf{Implementation details.}
VICR uses FLUX.1-Fill-dev~\citep{flux2024fill} as the generative backbone and is fine-tuned with LoRA~\citep{hu2022lora} using rank 32. LoRA adapters are applied to the FLUX transformer conditioning and attention-related projection layers, while global image context is extracted with CLIP ViT-L/14~\citep{radford2021clip} and adapted through the Vision-Context Bridge. Training uses \(512\times512\) HR crops, a batch size of 16, and the Prodigy optimizer~\citep{mishchenko2023prodigy} with bias correction, safeguard warmup, and a weight decay of 0.01. The model is trained on 8 NVIDIA H800 GPUs for 15K iterations, taking approximately 24 hours. During inference, we use FlowMatchEulerDiscreteScheduler~\citep{vonplaten2022diffusers} with 30 sampling steps and a guidance scale of 30.

\noindent\textbf{Evaluation metrics.}
Consistent with DiT4SR and prior studies~\citep{blau2018perception,gu2020pipal,yu2024supir,duan2025dit4sr}, we use LPIPS~\citep{zhang2018lpips} as the primary full-reference perceptual metric and report PSNR/SSIM~\citep{wang2004ssim} only as supplementary fidelity indicators, since these distortion-oriented scores often fail to reflect the visual quality of generative SR results. On paired benchmarks, we additionally report DISTS~\citep{ding2020dists} and FID~\citep{heusel2017fid}, with PSNR/SSIM results provided in Appendix~\ref{app:text_supervision_ablation}. On real images without reference targets, we report LIQE~\citep{zhang2023liqe}, MUSIQ~\citep{ke2021musiq}, MANIQA~\citep{yang2022maniqa}, and CLIPIQA~\citep{wang2023clipiqa}. All metrics use the same pyiqa backend~\citep{chen2022pyiqa}; protocol details are in Appendix~\ref{app:experimental_details}.

\noindent\textbf{Compared methods.}
We compare VICR with representative recent generative Real-ISR methods, including StableSR~\citep{wang2023stablesr}, SeeSR~\citep{wu2024seesr}, SUPIR~\citep{yu2024supir}, PASD~\citep{yang2024pasd}, DiffBIR~\citep{lin2024diffbir}, OSEDiff~\citep{wu2024osediff}, FaithDiff~\citep{chen2025faithdiff}, PiSA-SR~\citep{sun2025pisasr}, DiT4SR~\citep{duan2025dit4sr}, and FluxControlNet. We use official checkpoints and configurations whenever available, and re-evaluate all methods under the unified pyiqa backend for fair comparison.

\subsection{Comparison with existing state-of-the-art methods}

\subsubsection{Quantitative analysis}

Table~\ref{tab:main_results} reports representative paired-benchmark comparisons, while the representative no-reference comparison on RealLR200 and RealLQ250 is provided in Appendix~\ref{app:additional_quantitative_results} (Table~\ref{tab:main_noref}). Full comparisons with all evaluated baselines are provided in Appendix~\ref{app:additional_quantitative_results} (Tables~\ref{tab:main_results_full} and~\ref{tab:main_noref_full}). As shown in Table~\ref{tab:main_results}, VICR shows different strengths with and without agent-driven prompt refinement. VICR w/o Agent Ref. achieves strong full-reference metrics and FID, while the full VICR pipeline further improves no-reference perceptual metrics. Across RealSR, DIV2K-Val, and DRealSR, the two VICR variants obtain leading results on FID and no-reference metrics, indicating strong visual realism and perceptual quality.

% Requires \usepackage{booktabs,multirow,xcolor,graphicx}
\begin{table*}[t]
\centering
\small
\renewcommand{\arraystretch}{1.12}
\setlength{\tabcolsep}{4pt}
\caption{Representative quantitative comparison on RealSR, DIV2K-Val, and DRealSR under the unified pyiqa backend. Red and blue denote the best and second-best results, respectively. VICR denotes the full inference pipeline with agent-driven prompt refinement. VICR w/o Agent Ref. uses the initial LLaVA prompt and a single output.}
\label{tab:main_results}
\resizebox{\textwidth}{!}{%
\begin{tabular}{clccccccc}
\toprule
\multirow{2}{*}{Dataset} & \multirow{2}{*}{Method} & \multicolumn{7}{c}{Metric} \\
\cmidrule(lr){3-9}
& & LPIPS$\downarrow$ & DISTS$\downarrow$ & FID$\downarrow$ & LIQE$\uparrow$ & MUSIQ$\uparrow$ & MANIQA$\uparrow$ & CLIPIQA$\uparrow$ \\
\midrule

\multirow{8}{*}{RealSR} & StableSR & 0.3272 & 0.2454 & 134.89 & 3.5055 & 61.47 & 0.4634 & 0.5653 \\
 & SeeSR & \textcolor{blue}{0.3007} & \textcolor{blue}{0.2224} & 125.51 & 4.1365 & \textcolor{blue}{69.82} & 0.5437 & 0.6704 \\
 & DiffBIR & 0.4012 & 0.2603 & 151.33 & \textcolor{blue}{4.2846} & 67.94 & 0.4865 & 0.6776 \\
 & FaithDiff & \textcolor{red}{\textbf{0.2887}} & \textcolor{red}{\textbf{0.2114}} & \textcolor{blue}{108.89} & 3.9926 & 68.85 & 0.4680 & 0.6204 \\
 & DiT4SR & 0.3201 & 0.2258 & 120.12 & 3.9017 & 67.63 & 0.4582 & 0.6294 \\
 & FluxControlNet & 0.3555 & 0.2438 & 126.46 & 3.6057 & 64.29 & 0.4554 & 0.6394 \\
\cmidrule(lr){2-9}
 & VICR w/o Agent Ref. & 0.3323 & 0.2359 & \textcolor{red}{\textbf{108.24}} & 4.2739 & 69.71 & \textcolor{blue}{0.6070} & \textcolor{blue}{0.7007} \\
 & VICR & 0.3383 & 0.2383 & 110.46 & \textcolor{red}{\textbf{4.3833}} & \textcolor{red}{\textbf{70.52}} & \textcolor{red}{\textbf{0.6222}} & \textcolor{red}{\textbf{0.7155}} \\
\midrule

\multirow{8}{*}{DIV2K-Val} & StableSR & 0.3456 & 0.2142 & 31.04 & 4.1023 & 65.07 & 0.5393 & 0.7124 \\
 & SeeSR & 0.3190 & \textcolor{blue}{0.1963} & 26.02 & 4.2642 & 68.62 & 0.5030 & 0.6926 \\
 & DiffBIR & 0.3741 & 0.2291 & 42.09 & 4.3054 & 67.47 & 0.4685 & 0.6868 \\
 & FaithDiff & \textcolor{blue}{0.3124} & 0.1992 & 25.88 & 4.2441 & 69.32 & 0.4311 & 0.6467 \\
 & DiT4SR & 0.3427 & 0.2123 & 36.60 & 4.1329 & 67.39 & 0.4392 & 0.6602 \\
 & FluxControlNet & 0.3580 & 0.2137 & 26.51 & 3.7580 & 62.35 & 0.4321 & 0.6164 \\
\cmidrule(lr){2-9}
 & VICR w/o Agent Ref. & \textcolor{red}{\textbf{0.3109}} & \textcolor{red}{\textbf{0.1955}} & \textcolor{red}{\textbf{23.82}} & \textcolor{blue}{4.5183} & \textcolor{blue}{70.12} & \textcolor{blue}{0.6103} & \textcolor{blue}{0.7283} \\
 & VICR & 0.3179 & 0.1989 & \textcolor{blue}{24.93} & \textcolor{red}{\textbf{4.6074}} & \textcolor{red}{\textbf{70.98}} & \textcolor{red}{\textbf{0.6251}} & \textcolor{red}{\textbf{0.7431}} \\
\midrule

\multirow{8}{*}{DRealSR} & StableSR & 0.3792 & 0.2683 & 169.83 & 2.9169 & 53.03 & 0.3943 & 0.5328 \\
 & SeeSR & \textcolor{red}{\textbf{0.3174}} & \textcolor{red}{\textbf{0.2315}} & 147.46 & \textcolor{blue}{4.1273} & 65.09 & 0.5129 & \textcolor{red}{\textbf{0.6910}} \\
 & DiffBIR & 0.5095 & 0.2959 & 175.92 & 3.7407 & 59.83 & 0.4508 & 0.6486 \\
 & FaithDiff & \textcolor{blue}{0.3501} & \textcolor{blue}{0.2416} & 153.34 & 4.0124 & \textcolor{blue}{66.35} & 0.4512 & 0.6332 \\
 & DiT4SR & 0.3721 & 0.2445 & 156.99 & 3.9953 & 65.29 & 0.4439 & 0.6660 \\
 & FluxControlNet & 0.3972 & 0.2521 & 156.29 & 3.1211 & 56.01 & 0.4069 & 0.5959 \\
\cmidrule(lr){2-9}
 & VICR w/o Agent Ref. & 0.3662 & 0.2456 & \textcolor{red}{\textbf{138.92}} & 3.9471 & 64.14 & \textcolor{blue}{0.5163} & 0.6597 \\
 & VICR & 0.3863 & 0.2532 & \textcolor{blue}{146.34} & \textcolor{red}{\textbf{4.1324}} & \textcolor{red}{\textbf{66.39}} & \textcolor{red}{\textbf{0.5484}} & \textcolor{blue}{0.6874} \\
\bottomrule
\end{tabular}%
}
\end{table*}

On RealSR and DRealSR, VICR does not lead on the two full-reference metrics, LPIPS and DISTS\@. We attribute this mainly to the reference quality of these two benchmarks rather than to degraded visual quality. As shown by the GT scores in Appendix~\ref{app:benchmark_reference_quality} (Table~\ref{tab:gt_noref_scores}), the HR targets of RealSR and DRealSR receive relatively low no-reference quality scores, suggesting that some references are themselves blurry or less aligned with the naturalness preferred by perceptual IQA models. Therefore, results that are more visually natural and sharper may deviate from such references under full-reference metrics, while still being favored by FID and no-reference perceptual metrics. This behavior is also consistent with agent-driven prompt refinement: more specific and realistic synthesized details can improve no-reference perceptual scores, but these details may not exactly match the provided GT.

The no-reference results in Appendix~\ref{app:additional_quantitative_results} (Table~\ref{tab:main_noref}) further verify the generalization ability of VICR on real images without reference targets. On RealLR200, VICR achieves the best MANIQA and CLIPIQA scores and remains highly competitive on LIQE and MUSIQ\@. On RealLQ250, VICR achieves the best MUSIQ, MANIQA, and CLIPIQA scores, while ranking second on LIQE\@. Overall, VICR generalizes well beyond the main benchmarks and consistently produces more realistic and perceptually favorable Real-ISR results on diverse real-world images. VICR also uses only 127M trainable parameters, fewer than representative generative Real-ISR baselines; the detailed comparison is provided in Appendix~\ref{app:experimental_details}.

\subsubsection{Qualitative analysis}

\begin{figure*}[t]
    \centering
    \includegraphics[width=\textwidth]{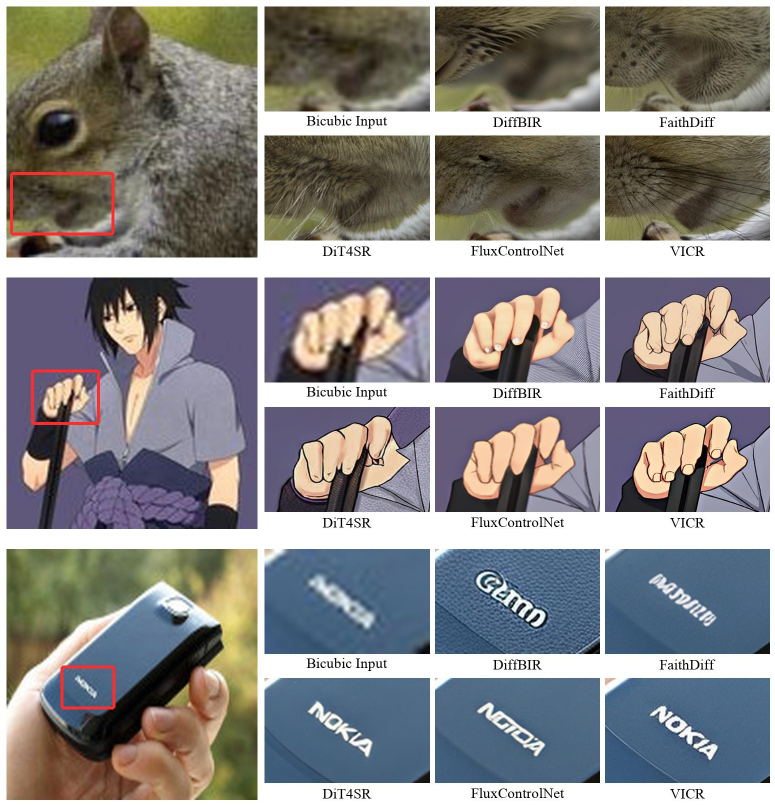}
    \caption{Qualitative comparison between VICR and representative Real-ISR methods on challenging low-quality images. Red boxes indicate enlarged regions. Existing methods often oversmooth textures or synthesize sharp but spatially inconsistent details under severe degradations. VICR coordinates local visual evidence, global image context, and agent-refined semantic prompts, restoring realistic fine details while preserving input structures and reducing hallucinated or misaligned content.}
    \label{fig:qualitative_comparison}
\end{figure*}

\noindent\textbf{Qualitative comparison.}
Figure~\ref{fig:qualitative_comparison} shows that existing methods often struggle to balance perceptual sharpness with structural fidelity under severe degradations. Some baselines oversmooth fine textures, while others synthesize sharper but misaligned details, such as distorted animal textures, unstable hand contours, or incorrect logo/text patterns. In contrast, VICR better follows the LQ structure while restoring realistic local details. Consistent with the quantitative results, the proposed condition organization improves visual realism while reducing spatially inconsistent hallucinations. More qualitative comparisons on diverse real-world inputs and enlarged local regions are provided in Appendix~\ref{app:additional_qualitative_results}.

\subsection{Ablation Studies and Analysis}

Beyond the main comparisons, we analyze VICR from four aspects: visual condition organization, agent-driven prompt refinement, text supervision, and parameter efficiency.

\begin{table*}[t]
\centering
\small
\setlength{\tabcolsep}{5pt}
\renewcommand{\arraystretch}{1.12}
\caption{Condition-organization ablation on the RealLR200 benchmark. All variants use the same static LLaVA caption to isolate the effect of visual condition organization; A2 adapts the DiT4SR-style token injection strategy to Flux Fill.}
\label{tab:ablation_vicr_reallr200}
\resizebox{\textwidth}{!}{%
\begin{tabular}{llcccccccc}
\toprule
& & \multicolumn{4}{c}{Condition organization} & \multicolumn{4}{c}{RealLR200} \\
\cmidrule(lr){3-6}\cmidrule(lr){7-10}
Variant & Name & Local evidence & Global context & VCB & Token inj. & MANIQA$\uparrow$ & MUSIQ$\uparrow$ & CLIPIQA$\uparrow$ & LIQE$\uparrow$ \\
\midrule
A1 & Local-only VICR                 & \checkmark & $\times$   & $\times$   & $\times$   & 0.5077 & 68.1042 & 0.6542 & 3.9563 \\
A2 & DiT4SR-style token injection    & $\times$   & $\times$   & $\times$   & \checkmark & 0.4067 & 61.1938 & 0.5678 & 3.1952 \\
A3 & Global-only VICR                & $\times$   & \checkmark & \checkmark & $\times$   & 0.5299 & \textbf{70.8224} & 0.6978 & \textbf{4.5649} \\
A4 & VICR w/o VCB                    & \checkmark & \checkmark & $\times$   & $\times$   & 0.4944 & 61.1463 & 0.6330 & 1.5596 \\
A5 & Full visual VICR                & \checkmark & \checkmark & \checkmark & $\times$   & \textbf{0.5674} & 69.8212 & \textbf{0.7144} & 4.2288 \\
\bottomrule
\end{tabular}%
}
\end{table*}

\noindent\textbf{Role of visual condition organization.}

Table~\ref{tab:ablation_vicr_reallr200} shows that the local-only variant already forms a meaningful SR baseline, while the DiT4SR-style token injection variant performs substantially worse, suggesting that direct visual-token injection is less compatible with Flux Fill than the proposed in-context formulation. The global-only variant obtains favorable no-reference scores but may drift from the LQ structure without local visual evidence, as visualized in Appendix~\ref{app:additional_ablations} (Figure~\ref{fig:appendix_global_only_fidelity}). The degraded performance of VICR w/o VCB further shows that global image context must be adapted before entering the DiT backbone. The full visual design achieves the best overall balance.

\noindent\textbf{Effect of agent-driven prompt refinement.}

Appendix~\ref{app:agent_prompt_ablation} evaluates agent-driven prompt refinement from raw outputs at each refinement round and Best-of-\(K\) post-selection. Later raw rounds improve over the static LLaVA prompt on all four no-reference metrics, and the Best-of-\(K\) curve from \(K=1\) to \(K=10\) shows an overall upward trend. These results demonstrate the effectiveness of inference-time agent refinement without updating model parameters.
We report raw round-wise outputs and Best-of-\(K\) results separately: the former evaluates agent refinement without post-selection, whereas the latter evaluates the selected output set obtained from the first \(K\) generated candidates.

\noindent\textbf{Effect of text supervision.}

Text supervision mainly changes the fidelity--perception tendency of VICR\@. Appendix~\ref{app:text_supervision_ablation} compares DAPE tags and LLaVA-generated image descriptions. DAPE tags provide weaker semantic guidance and favor GT-aligned, fidelity-oriented SR, whereas LLaVA captions provide richer scene context and favor perceptual SR under no-reference metrics. We use LLaVA captions for the main VICR model.

\noindent\textbf{Parameter efficiency.}

Appendix~\ref{app:experimental_details} shows that VICR uses only 127M trainable parameters, fewer than the listed generative Real-ISR baselines, while still achieving strong perceptual performance.

\section{Conclusion}

This paper studies condition organization for generative real-world image super-resolution and proposes VICR, which formulates Real-ISR as visual in-context completion. By separating local visual evidence, global image context, and semantic prompts into complementary condition streams, VICR improves perceptual quality and visual realism across multiple SR benchmarks while using only 127M trainable parameters. These results indicate that assigning clear roles to different conditions is effective for generative Real-ISR.

\noindent\textbf{Limitations and future work.}
VICR currently uses a fixed-resolution diptych input, which limits arbitrary-resolution inference and may weaken consistency across image tiles during tiled processing. Future work will explore more flexible diptych construction and consistency constraints across image tiles; more discussion is provided in Appendix~\ref{app:limitations}.

\bibliographystyle{unsrtnat}
\bibliography{references}

@article{dong2016srcnn,
  title={Image Super-Resolution Using Deep Convolutional Networks},
  author={Dong, Chao and Loy, Chen Change and He, Kaiming and Tang, Xiaoou},
  journal={IEEE Transactions on Pattern Analysis and Machine Intelligence},
  volume={38},
  number={2},
  pages={295--307},
  year={2016}
}

@inproceedings{lim2017edsr,
  title={Enhanced Deep Residual Networks for Single Image Super-Resolution},
  author={Lim, Bee and Son, Sanghyun and Kim, Heewon and Nah, Seungjun and Lee, Kyoung Mu},
  booktitle={Proceedings of the IEEE Conference on Computer Vision and Pattern Recognition Workshops},
  pages={136--144},
  year={2017}
}

@inproceedings{zhang2018rcan,
  title={Image Super-Resolution Using Very Deep Residual Channel Attention Networks},
  author={Zhang, Yulun and Li, Kunpeng and Li, Kai and Wang, Lichen and Zhong, Bineng and Fu, Yun},
  booktitle={Proceedings of the European Conference on Computer Vision},
  pages={286--301},
  year={2018}
}

@inproceedings{ledig2017srgan,
  title={Photo-Realistic Single Image Super-Resolution Using a Generative Adversarial Network},
  author={Ledig, Christian and Theis, Lucas and Huszar, Ferenc and Caballero, Jose and Cunningham, Andrew and Acosta, Alejandro and Aitken, Andrew and Tejani, Alykhan and Totz, Johannes and Wang, Zehan and Shi, Wenzhe},
  booktitle={Proceedings of the IEEE Conference on Computer Vision and Pattern Recognition},
  pages={4681--4690},
  year={2017}
}

@inproceedings{sajjadi2017enhancenet,
  title={{EnhanceNet}: Single Image Super-Resolution Through Automated Texture Synthesis},
  author={Sajjadi, Mehdi S. M. and Scholkopf, Bernhard and Hirsch, Michael},
  booktitle={Proceedings of the IEEE International Conference on Computer Vision},
  pages={4501--4510},
  year={2017}
}

@inproceedings{wang2018sftgan,
  title={Recovering Realistic Texture in Image Super-Resolution by Deep Spatial Feature Transform},
  author={Wang, Xintao and Yu, Ke and Dong, Chao and Loy, Chen Change},
  booktitle={Proceedings of the IEEE Conference on Computer Vision and Pattern Recognition},
  pages={606--615},
  year={2018}
}

@inproceedings{wang2018esrgan,
  title={{ESRGAN}: Enhanced Super-Resolution Generative Adversarial Networks},
  author={Wang, Xintao and Yu, Ke and Wu, Shixiang and Gu, Jinjin and Liu, Yihao and Dong, Chao and Qiao, Yu and Change Loy, Chen},
  booktitle={Proceedings of the European Conference on Computer Vision Workshops},
  pages={63--79},
  year={2018},
  doi={10.1007/978-3-030-11021-5_5}
}

@inproceedings{zhang2019ranksrgan,
  title={{RankSRGAN}: Generative Adversarial Networks with Ranker for Image Super-Resolution},
  author={Zhang, Wenlong and Liu, Yihao and Dong, Chao and Qiao, Yu},
  booktitle={Proceedings of the IEEE/CVF International Conference on Computer Vision},
  pages={3096--3105},
  year={2019}
}

@inproceedings{blau2018perception,
  title={The Perception-Distortion Tradeoff},
  author={Blau, Yochai and Michaeli, Tomer},
  booktitle={Proceedings of the IEEE Conference on Computer Vision and Pattern Recognition},
  pages={6228--6237},
  year={2018}
}

@inproceedings{gu2020pipal,
  title={{PIPAL}: A Large-Scale Image Quality Assessment Dataset for Perceptual Image Restoration},
  author={Gu, Jinjin and Cai, Haoming and Chen, Haoyu and Ye, Xiaoxing and Ren, Jimmy S. and Dong, Chao},
  booktitle={Proceedings of the European Conference on Computer Vision},
  pages={633--651},
  year={2020}
}

@inproceedings{fritsche2019frequency,
  title={Frequency Separation for Real-World Super-Resolution},
  author={Fritsche, Manuel and Gu, Shuhang and Timofte, Radu},
  booktitle={Proceedings of the IEEE/CVF International Conference on Computer Vision Workshops},
  pages={3599--3608},
  year={2019}
}

@inproceedings{maeda2020unpaired,
  title={Unpaired Image Super-Resolution Using Pseudo-Supervision},
  author={Maeda, Shunta},
  booktitle={Proceedings of the IEEE/CVF Conference on Computer Vision and Pattern Recognition},
  pages={291--300},
  year={2020}
}

@inproceedings{wang2021realesrgan,
  title={Real-{ESRGAN}: Training Real-World Blind Super-Resolution with Pure Synthetic Data},
  author={Wang, Xintao and Xie, Liangbin and Dong, Chao and Shan, Ying},
  booktitle={Proceedings of the IEEE/CVF International Conference on Computer Vision Workshops},
  pages={1905--1914},
  year={2021}
}

@inproceedings{zhang2021bsrgan,
  title={Designing a Practical Degradation Model for Deep Blind Image Super-Resolution},
  author={Zhang, Kai and Liang, Jingyun and Van Gool, Luc and Timofte, Radu},
  booktitle={Proceedings of the IEEE/CVF International Conference on Computer Vision},
  pages={4791--4800},
  year={2021}
}

@inproceedings{wei2021dasr,
  title={Unsupervised Real-World Image Super Resolution via Domain-Distance Aware Training},
  author={Wei, Yunxuan and Gu, Shuhang and Li, Yawei and Timofte, Radu and Jin, Longcun and Song, Hengjie},
  booktitle={Proceedings of the IEEE/CVF Conference on Computer Vision and Pattern Recognition},
  pages={13385--13394},
  year={2021}
}

@inproceedings{liang2021swinir,
  title={{SwinIR}: Image Restoration Using Swin Transformer},
  author={Liang, Jingyun and Cao, Jiezhang and Sun, Guolei and Zhang, Kai and Van Gool, Luc and Timofte, Radu},
  booktitle={Proceedings of the IEEE/CVF International Conference on Computer Vision Workshops},
  pages={1833--1844},
  year={2021}
}

@inproceedings{conde2022swin2sr,
  title={{Swin2SR}: SwinV2 Transformer for Compressed Image Super-Resolution and Restoration},
  author={Conde, Marcos V. and Choi, Ui-Jin and Burchi, Maxime and Timofte, Radu},
  booktitle={Proceedings of the European Conference on Computer Vision Workshops},
  pages={669--687},
  year={2022},
  doi={10.1007/978-3-031-25063-7_42}
}

@article{wang2023stablesr,
  title={Exploiting Diffusion Prior for Real-World Image Super-Resolution},
  author={Wang, Jianyi and Yue, Zongsheng and Zhou, Shangchen and Chan, Kelvin C. K. and Loy, Chen Change},
  journal={International Journal of Computer Vision},
  volume={132},
  number={12},
  pages={5929--5949},
  year={2024},
  doi={10.1007/s11263-024-02168-7},
  eprint={2305.07015},
  archivePrefix={arXiv},
  primaryClass={cs.CV}
}

@inproceedings{lin2024diffbir,
  title={{DiffBIR}: Toward Blind Image Restoration with Generative Diffusion Prior},
  author={Lin, Xinqi and He, Jingwen and Chen, Ziyan and Lyu, Zhaoyang and Dai, Bo and Yu, Fanghua and Qiao, Yu and Ouyang, Wanli and Dong, Chao},
  booktitle={Proceedings of the European Conference on Computer Vision},
  pages={430--448},
  year={2024}
}

@inproceedings{yang2024pasd,
  title={Pixel-Aware Stable Diffusion for Realistic Image Super-Resolution and Personalized Stylization},
  author={Yang, Tao and Wu, Rongyuan and Ren, Peiran and Xie, Xuansong and Zhang, Yabin},
  booktitle={Proceedings of the European Conference on Computer Vision},
  pages={74--91},
  year={2024}
}

@inproceedings{yu2024supir,
  title={Scaling Up to Excellence: Practicing Model Scaling for Photo-Realistic Image Restoration In the Wild},
  author={Yu, Fanghua and Gu, Jinjin and Li, Zheyuan and Hu, Jinfan and Kong, Xiangtao and Wang, Xintao and He, Jingwen and Qiao, Yu and Dong, Chao},
  booktitle={Proceedings of the IEEE/CVF Conference on Computer Vision and Pattern Recognition},
  pages={25669--25680},
  year={2024}
}

@inproceedings{chen2025faithdiff,
  title={{FaithDiff}: Unleashing Diffusion Priors for Faithful Image Super-resolution},
  author={Chen, Junyang and Pan, Jinshan and Dong, Jiangxin},
  booktitle={Proceedings of the IEEE/CVF Conference on Computer Vision and Pattern Recognition},
  pages={28188--28197},
  year={2025}
}

@inproceedings{wu2024seesr,
  title={{SeeSR}: Towards Semantics-Aware Real-World Image Super-Resolution},
  author={Wu, Rongyuan and Yang, Tao and Sun, Lingchen and Zhang, Zhengqiang and Li, Shuai and Zhang, Lei},
  booktitle={Proceedings of the IEEE/CVF Conference on Computer Vision and Pattern Recognition},
  pages={25456--25467},
  year={2024}
}

@misc{chen2023promptsr,
  title={Image Super-Resolution with Text Prompt Diffusion},
  author={Chen, Zheng and Zhang, Yulun and Gu, Jinjin and Yuan, Xin and Kong, Linghe and Chen, Guihai and Yang, Xiaokang},
  year={2023},
  eprint={2311.14282},
  archivePrefix={arXiv},
  primaryClass={cs.CV}
}

@misc{jiang2024dalpsr,
  title={{DaLPSR}: Leverage Degradation-Aligned Language Prompt for Real-World Image Super-Resolution},
  author={Jiang, Aiwen and Wei, Zhi and Peng, Long and Liu, Feiqiang and Li, Wenbo and Wang, Mingwen},
  year={2024},
  eprint={2406.16477},
  archivePrefix={arXiv},
  primaryClass={cs.CV}
}

@inproceedings{chen2025srsr,
  title={{SRSR}: Enhancing Semantic Accuracy in Real-World Image Super-Resolution with Spatially Re-Focused Text-Conditioning},
  author={Chen, Chen and Abdolshah, Majid and Shevchenko, Violetta and Li, Hongdong and Xu, Chang and Purkait, Pulak},
  booktitle={Advances in Neural Information Processing Systems},
  volume={38},
  year={2025},
  eprint={2510.22534},
  archivePrefix={arXiv},
  primaryClass={cs.CV}
}

@inproceedings{hu2025tadisr,
  title={Text-Aware Real-World Image Super-Resolution via Diffusion Model with Joint Segmentation Decoders},
  author={Hu, Qiming and Fan, Linlong and Luo, Yiyan and Yu, Yuhang and Guo, Xiaojie and Fan, Qingnan},
  booktitle={Advances in Neural Information Processing Systems},
  volume={38},
  year={2025},
  eprint={2506.04641},
  archivePrefix={arXiv},
  primaryClass={cs.CV}
}

@inproceedings{kim2025chainofzoom,
  title={Chain-of-Zoom: Extreme Super-Resolution via Scale Autoregression and Preference Alignment},
  author={Kim, Bryan Sangwoo and Kim, Jeongsol and Ye, Jong Chul},
  booktitle={Advances in Neural Information Processing Systems},
  volume={38},
  year={2025},
  eprint={2505.18600},
  archivePrefix={arXiv},
  primaryClass={cs.CV}
}

@inproceedings{yi2025tvt,
  title={Fine-Structure Preserved Real-World Image Super-Resolution via Transfer {VAE} Training},
  author={Yi, Qiaosi and Li, Shuai and Wu, Rongyuan and Sun, Lingchen and Wu, Yuhui and Zhang, Lei},
  booktitle={Proceedings of the IEEE/CVF International Conference on Computer Vision},
  pages={12415--12426},
  year={2025}
}

@inproceedings{duan2025dit4sr,
  title={{DiT4SR}: Taming Diffusion Transformer for Real-World Image Super-Resolution},
  author={Duan, Zheng-Peng and Zhang, Jiawei and Jin, Xin and Zhang, Ziheng and Xiong, Zheng and Zou, Dongqing and Ren, Jimmy S. and Guo, Chunle and Li, Chongyi},
  booktitle={Proceedings of the IEEE/CVF International Conference on Computer Vision},
  pages={18948--18958},
  year={2025},
  eprint={2503.23580},
  archivePrefix={arXiv},
  primaryClass={cs.CV}
}

@inproceedings{sun2025pisasr,
  title={Pixel-level and Semantic-level Adjustable Super-resolution: A Dual-{LoRA} Approach},
  author={Sun, Lingchen and Wu, Rongyuan and Ma, Zhiyuan and Liu, Shuaizheng and Yi, Qiaosi and Zhang, Lei},
  booktitle={Proceedings of the IEEE/CVF Conference on Computer Vision and Pattern Recognition},
  pages={2333--2343},
  year={2025}
}

@inproceedings{wu2024osediff,
  title={One-Step Effective Diffusion Network for Real-World Image Super-Resolution},
  author={Wu, Rongyuan and Sun, Lingchen and Ma, Zhiyuan and Zhang, Lei},
  booktitle={Advances in Neural Information Processing Systems},
  volume={37},
  year={2024}
}

@inproceedings{ai2024dreamclear,
  title={{DreamClear}: High-Capacity Real-World Image Restoration with Privacy-Safe Dataset Curation},
  author={Ai, Yuang and Zhou, Xiaoqiang and Huang, Huaibo and Han, Xiaotian and Chen, Zhengyu and You, Quanzeng and Yang, Hongxia},
  booktitle={Advances in Neural Information Processing Systems},
  volume={37},
  year={2024},
  eprint={2410.18666},
  archivePrefix={arXiv},
  primaryClass={cs.CV}
}

@inproceedings{kong2025dpir,
  title={Dual Prompting Image Restoration with Diffusion Transformers},
  author={Kong, Dehong and Li, Fan and Wang, Zhixin and Xu, Jiaqi and Pei, Renjing and Li, Wenbo and Ren, WenQi},
  booktitle={Proceedings of the IEEE/CVF Conference on Computer Vision and Pattern Recognition},
  pages={12809--12819},
  year={2025}
}

@inproceedings{rombach2022ldm,
  title={High-Resolution Image Synthesis with Latent Diffusion Models},
  author={Rombach, Robin and Blattmann, Andreas and Lorenz, Dominik and Esser, Patrick and Ommer, Bjorn},
  booktitle={Proceedings of the IEEE/CVF Conference on Computer Vision and Pattern Recognition},
  pages={10684--10695},
  year={2022}
}

@inproceedings{peebles2023dit,
  title={Scalable Diffusion Models with Transformers},
  author={Peebles, William and Xie, Saining},
  booktitle={Proceedings of the IEEE/CVF International Conference on Computer Vision},
  pages={4195--4205},
  year={2023}
}

@inproceedings{esser2024sd3,
  title={Scaling Rectified Flow Transformers for High-Resolution Image Synthesis},
  author={Esser, Patrick and Kulal, Sumith and Blattmann, Andreas and Entezari, Rahim and Muller, Jonas and Saini, Harry and Levi, Yam and Lorenz, Dominik and Sauer, Axel and Boesel, Frederic and Podell, Dustin and Dockhorn, Tim and English, Zion and Lacey, Kyle and Goodwin, Alex and Marek, Yannik and Rombach, Robin},
  booktitle={Proceedings of the 41st International Conference on Machine Learning},
  pages={12606--12633},
  year={2024}
}

@misc{flux2025kontext,
  title={{FLUX.1 Kontext}: Flow Matching for In-Context Image Generation and Editing in Latent Space},
  author={{Black Forest Labs} and Batifol, Stephen and Blattmann, Andreas and Boesel, Frederic and Consul, Saksham and Diagne, Cyril and Dockhorn, Tim and English, Jack and English, Zion and Esser, Patrick and Kulal, Sumith and Lacey, Kyle and Levi, Yam and Li, Cheng and Lorenz, Dominik and Muller, Jonas and Podell, Dustin and Rombach, Robin and Saini, Harry and Sauer, Axel and Smith, Luke},
  year={2025},
  eprint={2506.15742},
  archivePrefix={arXiv},
  primaryClass={cs.GR}
}

@inproceedings{radford2021clip,
  title={Learning Transferable Visual Models From Natural Language Supervision},
  author={Radford, Alec and Kim, Jong Wook and Hallacy, Chris and Ramesh, Aditya and Goh, Gabriel and Agarwal, Sandhini and Sastry, Girish and Askell, Amanda and Mishkin, Pamela and Clark, Jack and Krueger, Gretchen and Sutskever, Ilya},
  booktitle={Proceedings of the International Conference on Machine Learning},
  pages={8748--8763},
  year={2021}
}

@inproceedings{liu2023llava,
  title={Visual Instruction Tuning},
  author={Liu, Haotian and Li, Chunyuan and Wu, Qingyang and Lee, Yong Jae},
  booktitle={Advances in Neural Information Processing Systems},
  volume={36},
  pages={34892--34916},
  year={2023},
  eprint={2304.08485},
  archivePrefix={arXiv},
  primaryClass={cs.CV}
}

@inproceedings{hu2022lora,
  title={{LoRA}: Low-Rank Adaptation of Large Language Models},
  author={Hu, Edward J. and Shen, Yelong and Wallis, Phillip and Allen-Zhu, Zeyuan and Li, Yuanzhi and Wang, Shean and Wang, Lu and Chen, Weizhu},
  booktitle={International Conference on Learning Representations},
  year={2022}
}

@misc{mishchenko2023prodigy,
  title={Prodigy: An Expeditiously Adaptive Parameter-Free Learner},
  author={Mishchenko, Konstantin and Defazio, Aaron},
  year={2023},
  eprint={2306.06101},
  archivePrefix={arXiv},
  primaryClass={cs.LG}
}

@misc{vonplaten2022diffusers,
  title={Diffusers: State-of-the-Art Diffusion Models},
  author={von Platen, Patrick and Patil, Suraj and Lozhkov, Anton and Cuenca, Pedro and Lambert, Nathan and Rasul, Kashif and Davaadorj, Mishig and Nair, Dhruv and Paul, Sayak and Berman, William and Xu, Yiyi and Liu, Steven and Wolf, Thomas},
  year={2022},
  howpublished={\url{https://github.com/huggingface/diffusers}}
}

@inproceedings{li2023lsdir,
  title={{LSDIR}: A Large Scale Dataset for Image Restoration},
  author={Li, Yawei and Zhang, Kai and Liang, Jingyun and Cao, Jiezhang and Liu, Ce and Gong, Rui and Zhang, Yulun and Tang, Hao and Liu, Yun and Demandolx, Denis and Ranjan, Rakesh and Timofte, Radu and Van Gool, Luc},
  booktitle={Proceedings of the IEEE/CVF Conference on Computer Vision and Pattern Recognition Workshops},
  pages={1775--1787},
  year={2023}
}

@inproceedings{karras2019stylegan,
  title={A Style-Based Generator Architecture for Generative Adversarial Networks},
  author={Karras, Tero and Laine, Samuli and Aila, Timo},
  booktitle={Proceedings of the IEEE/CVF Conference on Computer Vision and Pattern Recognition},
  pages={4401--4410},
  year={2019}
}

@inproceedings{agustsson2017div2k,
  title={{NTIRE} 2017 Challenge on Single Image Super-Resolution: Dataset and Study},
  author={Agustsson, Eirikur and Timofte, Radu},
  booktitle={Proceedings of the IEEE Conference on Computer Vision and Pattern Recognition Workshops},
  pages={126--135},
  year={2017}
}

@inproceedings{cai2019realsr,
  title={Toward Real-World Single Image Super-Resolution: A New Benchmark and a New Model},
  author={Cai, Jianrui and Zeng, Hui and Yong, Hongwei and Cao, Zisheng and Zhang, Lei},
  booktitle={Proceedings of the IEEE/CVF International Conference on Computer Vision},
  pages={3086--3095},
  year={2019}
}

@inproceedings{wei2020drealsr,
  title={Component Divide-and-Conquer for Real-World Image Super-Resolution},
  author={Wei, Pengxu and Xie, Ziwei and Lu, Hannan and Zhan, Zongyuan and Ye, Qixiang and Zuo, Wangmeng and Lin, Liang},
  booktitle={Proceedings of the European Conference on Computer Vision},
  pages={101--117},
  year={2020}
}

@article{wang2004ssim,
  title={Image Quality Assessment: From Error Visibility to Structural Similarity},
  author={Wang, Zhou and Bovik, Alan C. and Sheikh, Hamid R. and Simoncelli, Eero P.},
  journal={IEEE Transactions on Image Processing},
  volume={13},
  number={4},
  pages={600--612},
  year={2004}
}

@inproceedings{zhang2018lpips,
  title={The Unreasonable Effectiveness of Deep Features as a Perceptual Metric},
  author={Zhang, Richard and Isola, Phillip and Efros, Alexei A. and Shechtman, Eli and Wang, Oliver},
  booktitle={Proceedings of the IEEE Conference on Computer Vision and Pattern Recognition},
  pages={586--595},
  year={2018}
}

@article{ding2020dists,
  title={Image Quality Assessment: Unifying Structure and Texture Similarity},
  author={Ding, Keyan and Ma, Kede and Wang, Shiqi and Simoncelli, Eero P.},
  journal={IEEE Transactions on Pattern Analysis and Machine Intelligence},
  volume={44},
  number={5},
  pages={2567--2581},
  year={2022}
}

@inproceedings{heusel2017fid,
  title={{GANs} Trained by a Two Time-Scale Update Rule Converge to a Local Nash Equilibrium},
  author={Heusel, Martin and Ramsauer, Hubert and Unterthiner, Thomas and Nessler, Bernhard and Hochreiter, Sepp},
  booktitle={Advances in Neural Information Processing Systems},
  volume={30},
  year={2017}
}

@inproceedings{zhang2023liqe,
  title={Blind Image Quality Assessment via Vision-Language Correspondence: A Multitask Learning Perspective},
  author={Zhang, Weixia and Zhai, Guangtao and Wei, Ying and Yang, Xiaokang and Ma, Kede},
  booktitle={Proceedings of the IEEE/CVF Conference on Computer Vision and Pattern Recognition},
  pages={14071--14081},
  year={2023}
}

@inproceedings{ke2021musiq,
  title={{MUSIQ}: Multi-Scale Image Quality Transformer},
  author={Ke, Junjie and Wang, Qifei and Wang, Yilin and Milanfar, Peyman and Yang, Feng},
  booktitle={Proceedings of the IEEE/CVF International Conference on Computer Vision},
  pages={5148--5157},
  year={2021}
}

@inproceedings{yang2022maniqa,
  title={{MANIQA}: Multi-Dimension Attention Network for No-Reference Image Quality Assessment},
  author={Yang, Sidi and Wu, Tianhe and Shi, Shuwei and Lao, Shanshan and Gong, Yuan and Cao, Mingdeng and Wang, Jiahao and Yang, Yujiu},
  booktitle={Proceedings of the IEEE/CVF Conference on Computer Vision and Pattern Recognition Workshops},
  pages={1191--1200},
  year={2022}
}

@inproceedings{wang2023clipiqa,
  title={Exploring {CLIP} for Assessing the Look and Feel of Images},
  author={Wang, Jianyi and Chan, Kelvin C. K. and Loy, Chen Change},
  booktitle={Proceedings of the AAAI Conference on Artificial Intelligence},
  volume={37},
  pages={2555--2563},
  year={2023}
}

@misc{chen2022pyiqa,
  title={{IQA-PyTorch}: PyTorch Toolbox for Image Quality Assessment},
  author={Chen, Chaofeng and Mo, Jiadi},
  year={2022},
  howpublished={\url{https://github.com/chaofengc/IQA-PyTorch}}
}

@misc{flux2024fill,
  title={{FLUX.1 Fill [dev]}: Inpainting and Outpainting with Rectified Flow Transformers},
  author={{Black Forest Labs}},
  year={2024},
  howpublished={\url{https://huggingface.co/black-forest-labs/FLUX.1-Fill-dev}}
}

@misc{huang2024iclora,
  title={In-Context {LoRA} for Diffusion Transformers},
  author={Huang, Lianghua and Wang, Wei and Wu, Zhi-Fan and Shi, Yupeng and Dou, Huanzhang and Liang, Chen and Feng, Yutong and Liu, Yu and Zhou, Jingren},
  year={2024},
  eprint={2410.23775},
  archivePrefix={arXiv},
  primaryClass={cs.CV}
}

@inproceedings{shin2024diptych,
  title={Large-Scale Text-to-Image Model with Inpainting is a Zero-Shot Subject-Driven Image Generator},
  author={Shin, Chaehun and Choi, Jooyoung and Kim, Heeseung and Yoon, Sungroh},
  booktitle={Proceedings of the IEEE/CVF Conference on Computer Vision and Pattern Recognition},
  pages={7986--7996},
  year={2025},
  eprint={2411.15466},
  archivePrefix={arXiv},
  primaryClass={cs.CV}
}

@inproceedings{zhang2025icedit,
  title={In-Context Edit: Enabling Instructional Image Editing with In-Context Generation in Large Scale Diffusion Transformer},
  author={Zhang, Zechuan and Xie, Ji and Lu, Yu and Yang, Zongxin and Yang, Yi},
  booktitle={Advances in Neural Information Processing Systems},
  volume={38},
  year={2025},
  eprint={2504.20690},
  archivePrefix={arXiv},
  primaryClass={cs.CV}
}

@inproceedings{li2025visualcloze,
  title={{VisualCloze}: A Universal Image Generation Framework via Visual In-Context Learning},
  author={Li, Zhong-Yu and Du, Ruoyi and Yan, Juncheng and Zhuo, Le and Li, Zhen and Gao, Peng and Ma, Zhanyu and Cheng, Ming-Ming},
  booktitle={Proceedings of the IEEE/CVF International Conference on Computer Vision},
  pages={18969--18979},
  year={2025}
}

@misc{song2025insertanything,
  title={Insert Anything: Image Insertion via In-Context Editing in {DiT}},
  author={Song, Wensong and Jiang, Hong and Yang, Zongxing and Quan, Ruijie and Yang, Yi},
  year={2025},
  eprint={2504.15009},
  archivePrefix={arXiv},
  primaryClass={cs.CV}
}

@inproceedings{li2025iccustom,
  title={{IC-Custom}: Diverse Image Customization via In-Context Learning},
  author={Li, Yaowei and Li, Xiaoyu and Zhang, Zhaoyang and Bian, Yuxuan and Liu, Gan and Li, Xinyuan and Xu, Jiale and Hu, Wenbo and Liu, Yating and Li, Lingen and Cai, Jing and Zou, Yuexian and He, Yancheng and Shan, Ying},
  booktitle={International Conference on Learning Representations},
  year={2026},
  eprint={2507.01926},
  archivePrefix={arXiv},
  primaryClass={cs.CV}
}

@misc{jiang2026dtpsr,
  title={Disentangled Textual Priors for Diffusion-based Image Super-Resolution},
  author={Jiang, Lei and Liu, Xin and Tong, Xinze and Li, Zhiliang and Liu, Jie and Tang, Jie and Wu, Gangshan},
  year={2026},
  eprint={2603.07430},
  archivePrefix={arXiv},
  primaryClass={cs.CV}
}

@misc{xu2025incontextbrush,
  title={In-Context Brush: Zero-shot Customized Subject Insertion with Context-Aware Latent Space Manipulation},
  author={Xu, Yu and Tang, Fan and Wu, You and Gao, Lin and Deussen, Oliver and Yan, Hongbin and Li, Jintao and Cao, Juan and Lee, Tong-Yee},
  year={2025},
  eprint={2505.20271},
  archivePrefix={arXiv},
  primaryClass={cs.CV}
}

\clearpage
\appendix
\section{Supplementary Material Overview}

This appendix is organized by function rather than by the order in which materials were added. Appendix~\ref{app:condition_modeling} expands the motivation for condition organization in generative Real-ISR\@. Appendix~\ref{app:experimental_details} details the implementation, evaluation protocol, and parameter efficiency. Appendix~\ref{app:additional_quantitative_results} reports full quantitative comparisons. Appendix~\ref{app:additional_ablations} provides additional ablations and analysis. Appendix~\ref{app:additional_qualitative_results} presents additional qualitative examples. Appendices~\ref{app:limitations} and~\ref{app:broader_impacts} discuss limitations, future work, and broader impacts.

\section{Additional Discussion on Condition Organization}
\label{app:condition_modeling}

\subsection{Condition-injection paradigms}

Figure~\ref{fig:related_paradigms} summarizes three common ways of injecting conditions into generative Real-ISR models. Latent-level additive conditioning injects degraded-image information as a correction to the generative latent, but this shared perturbation must simultaneously encode local geometry, degradation cues, and semantic compensation. Auxiliary visual-control branches provide stronger image guidance, yet they still route low-level structure, scene-level context, and high-level semantics through a coupled control pathway. More recent token-level multimodal conditioning is more compatible with DiT backbones, but without explicit role assignment, visual tokens and text tokens are still left to negotiate local fidelity and semantic plausibility implicitly. These paradigms can improve perceptual quality, but they do not specify which condition should preserve spatial correspondence, which should stabilize global appearance, and which should resolve semantic ambiguity. This motivates the decoupled condition organization in VICR, where local visual evidence, global image context, and semantic prompts are assigned separate functions before being injected into the DiT backbone.

\subsection{DiT-based condition organization}

Many earlier diffusion-based Real-ISR pipelines are built on U-Net-based latent diffusion architectures~\citep{rombach2022ldm,wang2023stablesr,lin2024diffbir,yang2024pasd,wu2024seesr}. Since these models generate in the compressed latent space of pre-trained autoencoders~\citep{rombach2022ldm}, limited information bandwidth and the reconstruction ceiling may weaken fine-grained structures and textures~\citep{yi2025tvt}. DiT-based generative models provide a different interface: heterogeneous conditions can interact at the token level through attention mechanisms~\citep{peebles2023dit,esser2024sd3}. This makes DiT a natural backbone for organizing local visual evidence, global image context, and semantic prompts as distinct condition streams. However, existing DiT-based SR methods such as DiT4SR~\citep{duan2025dit4sr} still adopt joint token-level multimodal conditioning, rather than explicitly decomposing conditions according to their roles in the SR process.

\subsection{Semantic prompting under severe degradations}

Complex real-world degradations also expose a semantic gap that cannot be reliably resolved by image-only conditions. When local structures are severely blurred, corrupted, or missing, the low-quality input may provide insufficient evidence for recognizing scene semantics, object categories, and fine-grained visual attributes. Recent prompt-based SR methods introduce text or semantic prompts to complement such missing information~\citep{chen2023promptsr,yang2024pasd,wu2024seesr,jiang2024dalpsr,chen2025srsr,hu2025tadisr,kim2025chainofzoom}. However, many of these prompts are static: they are derived from fixed templates or generated once from the degraded input. Under severe degradations, such prompts remain constrained by incomplete visual evidence and cannot be updated when intermediate SR results reveal missing, ambiguous, or hallucinated details. This motivates our inference-time agent-driven prompting strategy, which treats textual guidance as an adaptive semantic prior coordinated with visual conditions during generation.

\section{Implementation and Evaluation Protocol}
\label{app:experimental_details}

\subsection{Evaluation protocol}

All full-reference benchmarks follow the \(4\times\) SR protocol and use the LR/HR pairs preprocessed by StableSR~\citep{wang2023stablesr}. For DIV2K, RealSR, and DRealSR, full-reference perceptual metrics are computed against the provided HR targets, while no-reference metrics are computed directly on SR outputs. RealLR200 and RealLQ250 do not provide reference HR targets, so we report only no-reference image quality metrics on these two datasets.

\subsection{Training and inference settings}

Section~4.1 summarizes the main training and inference hyperparameters. Here we emphasize that the same configuration is used for all main benchmarks and ablation variants unless explicitly stated. We do not tune the sampling steps, guidance scale, or training schedule separately for individual datasets, which keeps the paired and no-reference evaluations controlled under a unified protocol.

\subsection{LoRA adaptation and Vision-Context Bridge}

VICR adapts FLUX.1-Fill-dev with rank-32 LoRA adapters while keeping the base generative backbone frozen. The adapters are applied to conditioning-, attention-, and MLP-related projection layers in the FLUX transformer. The Vision-Context Bridge maps CLIP ViT-L/14 visual features into the FLUX conditioning space and is trained jointly with the LoRA adapters. Exact module patterns and configuration files are provided in the released code.

\subsection{Trainable parameter comparison}

\begin{table*}[t]
\centering
\caption{Comparison of trainable parameters among representative generative Real-ISR methods.}
\label{tab:trainable_params}
\renewcommand{\arraystretch}{1.15}
\setlength{\tabcolsep}{4pt}
\resizebox{\textwidth}{!}{%
\begin{tabular}{lcccccccc}
\toprule
Method 
& StableSR 
& SeeSR 
& SUPIR 
& PASD 
& DiffBIR 
& FaithDiff 
& DiT4SR 
& \textbf{VICR} \\
\midrule
Params$\downarrow$ 
& 149.91M 
& 749.9M 
& 1.3B 
& 609.52M 
& 364M 
& 2.61B 
& 350.88M 
& \textbf{127M} \\
\bottomrule
\end{tabular}%
}
\end{table*}

In addition to SR quality, we compare the number of trainable parameters to evaluate trainable-parameter efficiency. Table~\ref{tab:trainable_params} shows that VICR uses only 127M trainable parameters, which is the smallest among the listed generative Real-ISR baselines. Despite this compact adaptation, VICR still delivers strong perceptual performance in the quantitative comparisons of Section~4.2. This result suggests that high-quality Real-ISR does not necessarily require large additional control branches or expensive full-model fine-tuning. Instead, VICR benefits from a parameter-efficient conditioning design: its local visual evidence is provided directly through the diptych, avoiding a separate encoder for local LQ features, while the lightweight Vision-Context Bridge together with LoRA adapts global image context into the DiT conditioning space with few trainable parameters. Overall, VICR achieves a favorable balance between Real-ISR quality and trainable adaptation cost.

\section{Additional Quantitative Results}
\label{app:additional_quantitative_results}

\subsection{Full paired-benchmark comparison}

% Requires \usepackage{booktabs,multirow,xcolor,graphicx}
\begin{table*}[t]
\centering
\small
\renewcommand{\arraystretch}{1.12}
\setlength{\tabcolsep}{4pt}
\caption{Full quantitative comparison on RealSR, DIV2K-Val, and DRealSR under the unified pyiqa backend. Red and blue denote the best and second-best results, respectively. VICR denotes the full inference pipeline with agent-driven prompt refinement. VICR w/o Agent Ref. uses the initial LLaVA prompt and a single output.}
\label{tab:main_results_full}
\resizebox{\textwidth}{!}{%
\begin{tabular}{clccccccc}
\toprule
\multirow{2}{*}{Dataset} & \multirow{2}{*}{Method} & \multicolumn{7}{c}{Metric} \\
\cmidrule(lr){3-9}
& & LPIPS$\downarrow$ & DISTS$\downarrow$ & FID$\downarrow$ & LIQE$\uparrow$ & MUSIQ$\uparrow$ & MANIQA$\uparrow$ & CLIPIQA$\uparrow$ \\
\midrule

\multirow{12}{*}{RealSR} & StableSR & 0.3272 & 0.2454 & 134.89 & 3.5055 & 61.47 & 0.4634 & 0.5653 \\
 & SeeSR & 0.3007 & 0.2224 & 125.51 & 4.1365 & \textcolor{blue}{69.82} & 0.5437 & 0.6704 \\
 & SUPIR & 0.4757 & 0.2657 & 136.05 & 3.2000 & 60.95 & 0.4319 & 0.5714 \\
 & PASD & \textcolor{red}{\textbf{0.2859}} & \textcolor{red}{\textbf{0.2085}} & 127.95 & 2.8736 & 58.20 & 0.3767 & 0.4479 \\
 & DiffBIR & 0.4012 & 0.2603 & 151.33 & \textcolor{blue}{4.2846} & 67.94 & 0.4865 & 0.6776 \\
 & OSEDiff & 0.3173 & 0.2363 & 126.05 & 3.9402 & 67.53 & 0.4853 & 0.6735 \\
 & FaithDiff & \textcolor{blue}{0.2887} & \textcolor{blue}{0.2114} & \textcolor{blue}{108.89} & 3.9926 & 68.85 & 0.4680 & 0.6204 \\
 & PiSA-SR & 0.3130 & 0.2267 & 141.92 & 1.6308 & 47.13 & 0.3049 & 0.3593 \\
 & DiT4SR & 0.3201 & 0.2258 & 120.12 & 3.9017 & 67.63 & 0.4582 & 0.6294 \\
 & FluxControlNet & 0.3555 & 0.2438 & 126.46 & 3.6057 & 64.29 & 0.4554 & 0.6394 \\
\cmidrule(lr){2-9}
 & VICR w/o Agent Ref. & 0.3323 & 0.2359 & \textcolor{red}{\textbf{108.24}} & 4.2739 & 69.71 & \textcolor{blue}{0.6070} & \textcolor{blue}{0.7007} \\
 & VICR & 0.3383 & 0.2383 & 110.46 & \textcolor{red}{\textbf{4.3833}} & \textcolor{red}{\textbf{70.52}} & \textcolor{red}{\textbf{0.6222}} & \textcolor{red}{\textbf{0.7155}} \\
\midrule

\multirow{12}{*}{DIV2K-Val} & StableSR & 0.3456 & 0.2142 & 31.04 & 4.1023 & 65.07 & 0.5393 & 0.7124 \\
 & SeeSR & 0.3190 & \textcolor{blue}{0.1963} & 26.02 & 4.2642 & 68.62 & 0.5030 & 0.6926 \\
 & SUPIR & 0.3887 & 0.2133 & 26.44 & 4.0395 & 65.83 & 0.4451 & 0.6455 \\
 & PASD & 0.3843 & 0.2287 & 35.28 & 3.5030 & 60.53 & 0.3891 & 0.5380 \\
 & DiffBIR & 0.3741 & 0.2291 & 42.09 & 4.3054 & 67.47 & 0.4685 & 0.6868 \\
 & OSEDiff & \textcolor{red}{\textbf{0.3046}} & 0.2129 & 26.81 & 3.7510 & 65.56 & 0.4425 & 0.6589 \\
 & FaithDiff & 0.3124 & 0.1992 & 25.88 & 4.2441 & 69.32 & 0.4311 & 0.6467 \\
 & PiSA-SR & 0.4891 & 0.2693 & 46.02 & 1.4901 & 40.50 & 0.2560 & 0.3389 \\
 & DiT4SR & 0.3427 & 0.2123 & 36.60 & 4.1329 & 67.39 & 0.4392 & 0.6602 \\
 & FluxControlNet & 0.3580 & 0.2137 & 26.51 & 3.7580 & 62.35 & 0.4321 & 0.6164 \\
\cmidrule(lr){2-9}
 & VICR w/o Agent Ref. & \textcolor{blue}{0.3109} & \textcolor{red}{\textbf{0.1955}} & \textcolor{red}{\textbf{23.82}} & \textcolor{blue}{4.5183} & \textcolor{blue}{70.12} & \textcolor{blue}{0.6103} & \textcolor{blue}{0.7283} \\
 & VICR & 0.3179 & 0.1989 & \textcolor{blue}{24.93} & \textcolor{red}{\textbf{4.6074}} & \textcolor{red}{\textbf{70.98}} & \textcolor{red}{\textbf{0.6251}} & \textcolor{red}{\textbf{0.7431}} \\
\midrule

\multirow{12}{*}{DRealSR} & StableSR & 0.3792 & 0.2683 & 169.83 & 2.9169 & 53.03 & 0.3943 & 0.5328 \\
 & SeeSR & \textcolor{blue}{0.3174} & 0.2315 & 147.46 & \textcolor{blue}{4.1273} & 65.09 & 0.5129 & \textcolor{blue}{0.6910} \\
 & SUPIR & 0.5662 & 0.2931 & 158.32 & 3.3557 & 58.95 & 0.4436 & 0.5962 \\
 & PASD & 0.3224 & \textcolor{red}{\textbf{0.2273}} & 164.80 & 2.4823 & 49.89 & 0.3508 & 0.4704 \\
 & DiffBIR & 0.5095 & 0.2959 & 175.92 & 3.7407 & 59.83 & 0.4508 & 0.6486 \\
 & OSEDiff & 0.3177 & 0.2365 & \textcolor{blue}{142.18} & 3.8654 & 63.55 & 0.4790 & \textcolor{red}{\textbf{0.7060}} \\
 & FaithDiff & 0.3501 & 0.2416 & 153.34 & 4.0124 & \textcolor{blue}{66.35} & 0.4512 & 0.6332 \\
 & PiSA-SR & \textcolor{red}{\textbf{0.3054}} & \textcolor{blue}{0.2285} & 156.64 & 1.5760 & 43.53 & 0.3038 & 0.3680 \\
 & DiT4SR & 0.3721 & 0.2445 & 156.99 & 3.9953 & 65.29 & 0.4439 & 0.6660 \\
 & FluxControlNet & 0.3972 & 0.2521 & 156.29 & 3.1211 & 56.01 & 0.4069 & 0.5959 \\
\cmidrule(lr){2-9}
 & VICR w/o Agent Ref. & 0.3662 & 0.2456 & \textcolor{red}{\textbf{138.92}} & 3.9471 & 64.14 & \textcolor{blue}{0.5163} & 0.6597 \\
 & VICR & 0.3863 & 0.2532 & 146.34 & \textcolor{red}{\textbf{4.1324}} & \textcolor{red}{\textbf{66.39}} & \textcolor{red}{\textbf{0.5484}} & 0.6874 \\
\bottomrule
\end{tabular}%
}
\end{table*}

Table~\ref{tab:main_results_full} reports the full paired-benchmark comparison on DIV2K-Val, RealSR, and DRealSR under the unified pyiqa backend. As the complete version of Table~\ref{tab:main_results}, it confirms the same overall trend: VICR w/o Agent Ref. achieves strong FID and full-reference scores, while the full VICR pipeline obtains leading results on most no-reference metrics across the three paired benchmarks. The lower LPIPS/DISTS rankings on the two real-world paired benchmarks are consistent with the analysis in Section~4.2.

\subsection{Representative no-reference comparison}

\begin{table*}[t]
\centering
\small
\renewcommand{\arraystretch}{1.12}
\setlength{\tabcolsep}{3.5pt}
\caption{Representative no-reference quantitative comparison on RealLR200 and RealLQ250 under the unified pyiqa backend. Red and blue denote the best and second-best results, respectively.}
\label{tab:main_noref}
\resizebox{\textwidth}{!}{%
\begin{tabular}{lcccccccc}
\toprule
\multirow{2}{*}{Method}
& \multicolumn{4}{c}{RealLR200}
& \multicolumn{4}{c}{RealLQ250} \\
\cmidrule(lr){2-5}\cmidrule(lr){6-9}
& LIQE$\uparrow$ & MUSIQ$\uparrow$ & MANIQA$\uparrow$ & CLIPIQA$\uparrow$
& LIQE$\uparrow$ & MUSIQ$\uparrow$ & MANIQA$\uparrow$ & CLIPIQA$\uparrow$ \\
\midrule
StableSR
& 3.6087 & 64.53 & 0.4610 & 0.6540
& 3.5437 & 64.23 & 0.4279 & 0.6647 \\
SeeSR
& 4.0390 & 69.60 & 0.4977 & 0.6834
& 3.9787 & 70.38 & 0.4895 & 0.7061 \\
DiffBIR
& 3.8747 & 66.95 & 0.4576 & 0.6684
& 3.8728 & 66.68 & 0.4406 & 0.6703 \\
FaithDiff
& 3.9225 & 68.99 & 0.4285 & 0.6592
& 3.8442 & 69.70 & 0.3984 & 0.6652 \\
DiT4SR
& 4.1422 & 69.17 & 0.4552 & 0.6896
& 4.2885 & \textcolor{blue}{71.06} & 0.4593 & \textcolor{blue}{0.7120} \\
FluxControlNet
& \textcolor{red}{\textbf{4.4060}} & \textcolor{red}{\textbf{71.32}} & \textcolor{blue}{0.5186} & \textcolor{blue}{0.7038}
& \textcolor{red}{\textbf{4.3680}} & 70.58 & \textcolor{blue}{0.5085} & 0.7081 \\
VICR
& \textcolor{blue}{4.2898} & \textcolor{blue}{70.2334} & \textcolor{red}{\textbf{0.5770}} & \textcolor{red}{\textbf{0.7251}}
& \textcolor{blue}{4.3016} & \textcolor{red}{\textbf{71.0707}} & \textcolor{red}{\textbf{0.5449}} & \textcolor{red}{\textbf{0.7268}} \\
\bottomrule
\end{tabular}%
}
\end{table*}

Table~\ref{tab:main_noref} reports the representative no-reference comparison on RealLR200 and RealLQ250 discussed in the main paper. VICR obtains the best MANIQA and CLIPIQA scores on RealLR200 and the best MUSIQ, MANIQA, and CLIPIQA scores on RealLQ250, showing strong perceptual quality on real images without reference targets.

\subsection{Full no-reference comparison}

\begin{table*}[t]
\centering
\small
\renewcommand{\arraystretch}{1.12}
\setlength{\tabcolsep}{3.5pt}
\caption{Full no-reference quantitative comparison on RealLR200 and RealLQ250 under the unified pyiqa backend. Red and blue denote the best and second-best results, respectively.}
\label{tab:main_noref_full}
\resizebox{\textwidth}{!}{%
\begin{tabular}{lcccccccc}
\toprule
\multirow{2}{*}{Method}
& \multicolumn{4}{c}{RealLR200}
& \multicolumn{4}{c}{RealLQ250} \\
\cmidrule(lr){2-5}\cmidrule(lr){6-9}
& LIQE$\uparrow$ & MUSIQ$\uparrow$ & MANIQA$\uparrow$ & CLIPIQA$\uparrow$
& LIQE$\uparrow$ & MUSIQ$\uparrow$ & MANIQA$\uparrow$ & CLIPIQA$\uparrow$ \\
\midrule
StableSR
& 3.6087 & 64.53 & 0.4610 & 0.6540
& 3.5437 & 64.23 & 0.4279 & 0.6647 \\
SeeSR
& 4.0390 & 69.60 & 0.4977 & 0.6834
& 3.9787 & 70.38 & 0.4895 & 0.7061 \\
SUPIR
& 3.5202 & 64.15 & 0.3992 & 0.5573
& 3.6044 & 65.68 & 0.3825 & 0.5750 \\
PASD
& 3.1836 & 61.95 & 0.3767 & 0.5536
& 3.0007 & 60.62 & 0.3516 & 0.5397 \\
DiffBIR
& 3.8747 & 66.95 & 0.4576 & 0.6684
& 3.8728 & 66.68 & 0.4406 & 0.6703 \\
OSEDiff
& 3.7757 & 68.05 & 0.4452 & 0.6889
& 3.6201 & 67.59 & 0.4307 & 0.6911 \\
FaithDiff
& 3.9225 & 68.99 & 0.4285 & 0.6592
& 3.8442 & 69.70 & 0.3984 & 0.6652 \\
PiSA-SR
& 1.7138 & 47.86 & 0.2873 & 0.4444
& 1.3716 & 45.90 & 0.2583 & 0.3805 \\
DiT4SR
& 4.1422 & 69.17 & 0.4552 & 0.6896
& 4.2885 & \textcolor{blue}{71.06} & 0.4593 & \textcolor{blue}{0.7120} \\
FluxControlNet
& \textcolor{red}{\textbf{4.4060}} & \textcolor{red}{\textbf{71.32}} & \textcolor{blue}{0.5186} & \textcolor{blue}{0.7038}
& \textcolor{red}{\textbf{4.3680}} & 70.58 & \textcolor{blue}{0.5085} & 0.7081 \\
VICR
& \textcolor{blue}{4.2898} & \textcolor{blue}{70.2334} & \textcolor{red}{\textbf{0.5770}} & \textcolor{red}{\textbf{0.7251}}
& \textcolor{blue}{4.3016} & \textcolor{red}{\textbf{71.0707}} & \textcolor{red}{\textbf{0.5449}} & \textcolor{red}{\textbf{0.7268}} \\
\bottomrule
\end{tabular}%
}
\end{table*}

Table~\ref{tab:main_noref_full} reports the full no-reference comparison on RealLR200 and RealLQ250. Including all evaluated baselines does not change the main conclusion: VICR obtains the best MANIQA and CLIPIQA scores on RealLR200 and the best MUSIQ, MANIQA, and CLIPIQA scores on RealLQ250, while ranking second on LIQE on both datasets.

\section{Additional Ablations and Analysis}
\label{app:additional_ablations}

\subsection{Text supervision ablation}
\label{app:text_supervision_ablation}

In this paper, we examine two types of text supervision sources: one is DAPE tags following SeeSR~\citep{wu2024seesr}, and the other is natural language descriptions generated by LLaVA~\citep{liu2023llava} for the input low-quality images. Although both are used as text conditions for the generative model, they provide substantially different amounts of semantic information, which in turn affects how the model balances textual and visual conditions during SR generation and ultimately leads to different SR tendencies.

% Requires \usepackage{booktabs,multirow,xcolor,graphicx}
\begin{table*}[t]
\centering
\small
\renewcommand{\arraystretch}{1.12}
\setlength{\tabcolsep}{4pt}
\caption{Ablation of text supervision on RealSR, DIV2K-Val, and DRealSR\@. Red and blue denote the best and second-best results, respectively.}
\label{tab:prompt_ablation}
\resizebox{\textwidth}{!}{%
\begin{tabular}{clccccccccc}
\toprule
\multirow{2}{*}{Dataset} & \multirow{2}{*}{Method} & \multicolumn{9}{c}{Metric} \\
\cmidrule(lr){3-11}
& & PSNR$\uparrow$ & SSIM$\uparrow$ & LPIPS$\downarrow$ & DISTS$\downarrow$ & FID$\downarrow$ & LIQE$\uparrow$ & MUSIQ$\uparrow$ & MANIQA$\uparrow$ & CLIPIQA$\uparrow$ \\
\midrule

\multirow{3}{*}{RealSR}
 & VICR-Faithful    & \textcolor{red}{\textbf{24.30}} & \textcolor{red}{\textbf{0.7213}} & \textcolor{red}{\textbf{0.2577}} & \textcolor{red}{\textbf{0.2025}} & \textcolor{red}{\textbf{101.48}} & 3.4044 & 62.78 & 0.4365 & 0.5046 \\
 & VICR-Perceptual  & 22.41 & 0.6395 & 0.3326 & 0.2359 & \textcolor{blue}{108.16} & \textcolor{blue}{4.2715} & \textcolor{blue}{69.70} & \textcolor{blue}{0.6070} & \textcolor{blue}{0.7019} \\
 & VICR             & \textcolor{blue}{22.44} & \textcolor{blue}{0.6404} & \textcolor{blue}{0.3310} & \textcolor{blue}{0.2348} & 108.76 & \textcolor{red}{\textbf{4.3050}} & \textcolor{red}{\textbf{70.10}} & \textcolor{red}{\textbf{0.6110}} & \textcolor{red}{\textbf{0.7057}} \\
\midrule

\multirow{3}{*}{DIV2K-Val}
 & VICR-Faithful    & \textcolor{red}{\textbf{22.75}} & \textcolor{red}{\textbf{0.5748}} & \textcolor{red}{\textbf{0.2837}} & \textcolor{red}{\textbf{0.1827}} & \textcolor{red}{\textbf{21.39}} & 4.0781 & 65.70 & 0.5089 & 0.6134 \\
 & VICR-Perceptual  & 21.75 & 0.5313 & 0.3110 & 0.1955 & 23.81 & \textcolor{blue}{4.5179} & \textcolor{blue}{70.12} & \textcolor{blue}{0.6104} & \textcolor{blue}{0.7285} \\
 & VICR             & \textcolor{blue}{21.81} & \textcolor{blue}{0.5356} & \textcolor{blue}{0.3024} & \textcolor{blue}{0.1902} & \textcolor{blue}{22.76} & \textcolor{red}{\textbf{4.5396}} & \textcolor{red}{\textbf{70.23}} & \textcolor{red}{\textbf{0.6113}} & \textcolor{red}{\textbf{0.7296}} \\
\midrule

\multirow{3}{*}{DRealSR}
 & VICR-Faithful    & \textcolor{red}{\textbf{27.04}} & \textcolor{red}{\textbf{0.7463}} & \textcolor{red}{\textbf{0.2999}} & \textcolor{red}{\textbf{0.2148}} & \textcolor{red}{\textbf{129.19}} & 3.0211 & 56.24 & 0.3698 & 0.4629 \\
 & VICR-Perceptual  & \textcolor{blue}{25.20} & \textcolor{blue}{0.6722} & \textcolor{blue}{0.3665} & \textcolor{blue}{0.2456} & 138.88 & \textcolor{blue}{3.9460} & \textcolor{blue}{64.16} & \textcolor{blue}{0.5167} & \textcolor{blue}{0.6591} \\
 & VICR             & 24.99 & 0.6672 & 0.3775 & 0.2517 & \textcolor{blue}{131.28} & \textcolor{red}{\textbf{4.1533}} & \textcolor{red}{\textbf{66.63}} & \textcolor{red}{\textbf{0.5474}} & \textcolor{red}{\textbf{0.7088}} \\
\bottomrule
\end{tabular}%
}
\end{table*}

As shown in Table~\ref{tab:prompt_ablation}, VICR-Faithful trained with DAPE tags is overall more advantageous on full-reference metrics such as PSNR, SSIM, LPIPS, DISTS, and FID, whereas VICR-Perceptual trained with LLaVA captions performs better on no-reference metrics such as LIQE, MUSIQ, MANIQA, and CLIPIQA\@. A key reason is that DAPE tags provide much less semantic information than sentence-level captions. Under such weaker textual conditioning, the model is forced to rely more heavily on the image condition from the input LQ image itself, and the resulting outputs are therefore more likely to remain close to the reference GT\@. As a result, this setting tends to obtain better full-reference scores, which directly measure similarity to GT\@. In contrast, LLaVA captions provide richer scene context, object relations, and attribute information, and thus more strongly activate the high-level semantic and natural-image priors in the pretrained generative model. This makes the generated SR results more natural, sharper, and visually appealing, leading to stronger no-reference performance.

\subsection{Benchmark-reference quality analysis}
\label{app:benchmark_reference_quality}

\begin{table}[t]
\centering
\small
\renewcommand{\arraystretch}{1.10}
\setlength{\tabcolsep}{5pt}
\caption{No-reference quality scores of HR targets on the paired evaluation benchmarks. Higher values indicate better perceptual quality.}
\label{tab:gt_noref_scores}
\begin{tabular}{lcccc}
\toprule
Dataset & LIQE$\uparrow$ & MUSIQ$\uparrow$ & MANIQA$\uparrow$ & CLIPIQA$\uparrow$ \\
\midrule
RealSR & 2.6720 & 57.46 & 0.3295 & 0.4591 \\
DIV2K-Val & 3.8829 & 63.76 & 0.4060 & 0.6014 \\
DRealSR & 2.1154 & 50.56 & 0.3152 & 0.4750 \\
\bottomrule
\end{tabular}
\end{table}

The GT scores in Table~\ref{tab:gt_noref_scores} further reveal an important property of the RealSR and DRealSR benchmarks. On these two datasets, the GT images themselves receive relatively low no-reference scores on LIQE, MUSIQ, MANIQA, and CLIPIQA\@. Visual inspection further suggests that some GT images in these datasets are themselves relatively blurry and do not fully align with the high-naturalness distribution preferred by NR-IQA models. Therefore, on RealSR and DRealSR, a result that is more similar to GT may naturally obtain lower no-reference scores, while a result that is visually sharper and more natural may deviate from the relatively blurry GT and thus fail to achieve the best full-reference scores. In this sense, the observed fidelity--perception discrepancy is not caused only by the form of text supervision, but is also closely related to the properties of the benchmark GT itself.

This interpretation also explains why the conflict is weaker on DIV2K-Val. Since DIV2K-Val is constructed by synthesizing degradations from high-quality images, its GT images are themselves cleaner and closer to the natural image distribution preferred by generative models and NR-IQA metrics. Under this setting, full-reference and no-reference metrics are less contradictory than on RealSR and DRealSR\@. As a result, it becomes easier to obtain favorable performance on both types of metrics simultaneously. Overall, these results suggest that DAPE tags are better suited to fidelity-oriented SR because they provide less semantic information and force the model to rely more on the LQ input, whereas LLaVA captions are better suited to perceptual SR because they more strongly activate generative priors toward natural and visually pleasing outputs.

\subsection{Agent prompt refinement ablation}
\label{app:agent_prompt_ablation}

\paragraph{Agent instruction template.}
We summarize the fixed GPT-5.4 instruction template used for LLaVA-anchored closed-loop prompt refinement below.

\begin{center}
\setlength{\fboxsep}{6pt}
\noindent\fbox{%
\begin{minipage}{0.94\textwidth}
\small
\textbf{Inputs:} LQ image; initial LLaVA caption; previous-round raw SR result for refinement rounds; previous prompt.\\
\textbf{Initialization:} Use the LLaVA-generated caption as the initial semantic prompt.\\
\textbf{Refinement:} Edit the previous prompt conservatively rather than rewrite it, preserving the main scene description and adding only one or two cues about visible artifacts or ambiguous regions.\\
\textbf{Constraint:} No external web search or retrieved images are used.\\
\textbf{State rule:} Each refinement step uses only the raw previous-round output; best-of-previous-rounds, selected, oracle, or fallback outputs are not used as the next-step state.\\
\textbf{Constraints:} Do not invent object identities, readable text, or attributes unsupported by the LQ image.
\end{minipage}%
}
\end{center}

\paragraph{Example agent refinement.}
The following example shows that the agent makes a conservative full-prompt edit rather than a free-form rewrite.

\begin{center}
\setlength{\fboxsep}{6pt}
\noindent\fbox{%
\begin{minipage}{0.94\textwidth}
\small
\textbf{Initial prompt:} \emph{The image features a tree with no leaves, standing tall against a cloudy sky. The tree appears to be bare, with no foliage visible. The sky is overcast, creating a moody atmosphere in the scene.}\\[2pt]
\textbf{Previous prompt:} \emph{A bare winter tree line rises from the lower frame into a pale overcast sky, with dark left-leaning trunks and many fine leafless branches fading into light haze. The previous SR result is clear but still a little etched in twig tips and lower branch clusters, so refine gently: keep the sky smooth, cool, and low contrast, preserve the separated crowns, and soften black halos, duplicate twig strokes, crunchy sharpening, and small speckle around the upper branches.}\\[2pt]
\textbf{Next-round prompt:} \emph{A bare winter tree line rises from the lower frame into a pale overcast sky, with dark left-leaning trunks and many fine leafless branches fading into light haze. The current SR result is clear but still a little too black and wiry in twig tips and lower branch clusters, so refine gently: keep the sky smooth, cool, and low contrast, preserve separated crowns, and soften halos, duplicate twig strokes, crunchy sharpening, and speckle around upper branches.}
\end{minipage}%
}
\end{center}

\begin{table}[t]
\centering
\small
\setlength{\tabcolsep}{7pt}
\renewcommand{\arraystretch}{1.12}
\caption{Raw round-wise ablation of inference-time agent-driven prompt refinement on RealLR200. Static LLaVA initialization is the adopted round-1 input because it performs better than agent initialization, which is reported for comparison. Later rows report raw outputs from subsequent refinement rounds without Best-of-\(K\) selection.}
\label{tab:agent_refine_single_round_reallr200}
\resizebox{\columnwidth}{!}{%
\begin{tabular}{lcccc}
\toprule
Prompt setting & MANIQA$\uparrow$ & MUSIQ$\uparrow$ & CLIPIQA$\uparrow$ & LIQE$\uparrow$ \\
\midrule
Static LLaVA initialization (round 1) & 0.5683 & 69.8056 & 0.7152 & 4.2430 \\
Agent initialization (not adopted) & 0.5646 & 69.6605 & 0.7121 & 4.2012 \\
Agent refinement, round 2 & 0.5679 & 69.7336 & 0.7144 & 4.2326 \\
Agent refinement, round 3 & 0.5682 & 69.7769 & 0.7152 & 4.2362 \\
Agent refinement, round 4 & 0.5684 & 69.7803 & 0.7166 & 4.2383 \\
Agent refinement, round 5 & 0.5689 & 69.8128 & 0.7172 & 4.2403 \\
Agent refinement, round 6 & 0.5698 & 69.8177 & 0.7172 & 4.2421 \\
Agent refinement, round 7 & 0.5706 & 69.8331 & 0.7182 & 4.2436 \\
Agent refinement, round 8 & 0.5716 & 69.8828 & \textbf{0.7194} & 4.2490 \\
Agent refinement, round 9 & 0.5716 & 69.9195 & 0.7189 & 4.2520 \\
Agent refinement, round 10 & \textbf{0.5726} & \textbf{69.9498} & 0.7190 & \textbf{4.2586} \\
\bottomrule
\end{tabular}%
}
\end{table}

\begin{figure}[t]
    \centering
    \includegraphics[width=0.9\textwidth]{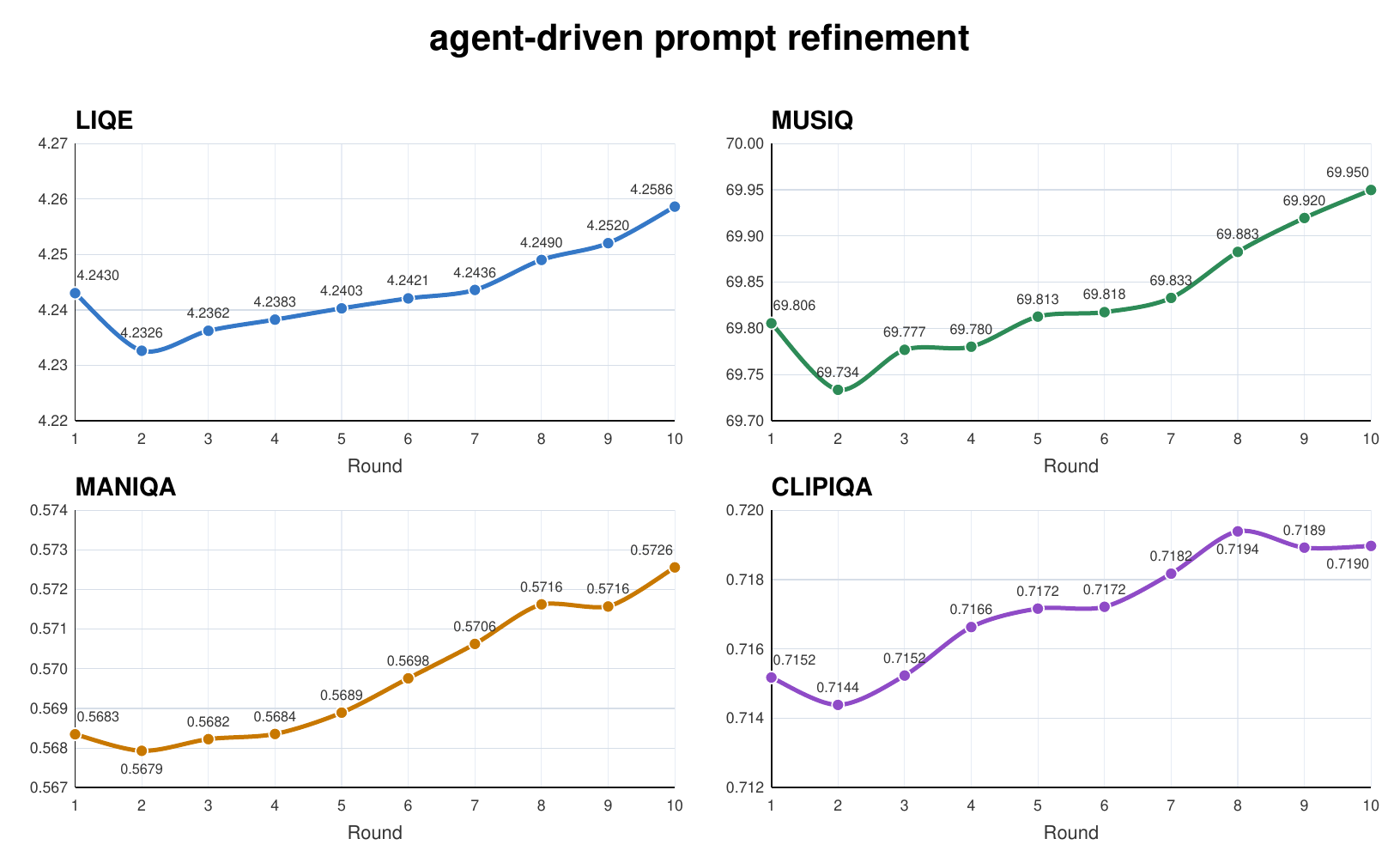}
    \caption{No-reference metric trends for raw outputs at each refinement round on RealLR200. Round 1 is the adopted static LLaVA initialization; no Best-of-\(K\) selection is applied to later raw outputs.}
    \label{fig:appendix_agent_raw_round_trend}
\end{figure}

Table~\ref{tab:agent_refine_single_round_reallr200} and Figure~\ref{fig:appendix_agent_raw_round_trend} report the adopted static LLaVA initialization and subsequent raw refinement rounds without Best-of-\(K\) selection. We also include the agent-initialized output in the table for comparison; it is weaker than the static LLaVA initialization, which motivates using the latter as the adopted starting point. The trend is not strictly monotonic in early rounds, but later raw refinement outputs overtake the adopted initialization on all four no-reference metrics.

\begin{table}[t]
\centering
\small
\setlength{\tabcolsep}{7pt}
\renewcommand{\arraystretch}{1.12}
\caption{Ablation of inference-time agent-driven prompt refinement on RealLR200. Static LLaVA initialization is used as \(K=1\) because it performs better than agent initialization, which is reported for comparison. For \(K \ge 2\), Best-of-\(K\) selects one output per image using the rank-sum rule over LIQE, MUSIQ, MANIQA, and CLIPIQA, then recomputes aggregate metrics on the selected set.}
\label{tab:agent_prompt_ablation}
\resizebox{\columnwidth}{!}{%
\begin{tabular}{lcccc}
\toprule
Prompt setting & MANIQA$\uparrow$ & MUSIQ$\uparrow$ & CLIPIQA$\uparrow$ & LIQE$\uparrow$ \\
\midrule
Static LLaVA initialization, \(K=1\) & 0.5683 & 69.8056 & 0.7152 & 4.2430 \\
Agent initialization (not adopted) & 0.5646 & 69.6605 & 0.7121 & 4.2012 \\
Agent refinement, \(K=2\) & 0.5715 & 69.9638 & 0.7178 & 4.2591 \\
Agent refinement, \(K=3\) & 0.5727 & 70.0562 & 0.7194 & 4.2629 \\
Agent refinement, \(K=4\) & 0.5732 & 70.0989 & 0.7206 & 4.2657 \\
Agent refinement, \(K=5\) & 0.5737 & 70.1191 & 0.7213 & 4.2675 \\
Agent refinement, \(K=6\) & 0.5743 & 70.1445 & 0.7218 & 4.2728 \\
Agent refinement, \(K=7\) & 0.5749 & 70.1604 & 0.7230 & 4.2767 \\
Agent refinement, \(K=8\) & 0.5760 & 70.2035 & 0.7245 & 4.2816 \\
Agent refinement, \(K=9\) & 0.5766 & 70.2015 & 0.7249 & 4.2857 \\
Agent refinement, \(K=10\) & \textbf{0.5770} & \textbf{70.2334} & \textbf{0.7251} & \textbf{4.2898} \\
\bottomrule
\end{tabular}%
}
\end{table}

\begin{figure}[t]
    \centering
    \includegraphics[width=0.9\textwidth]{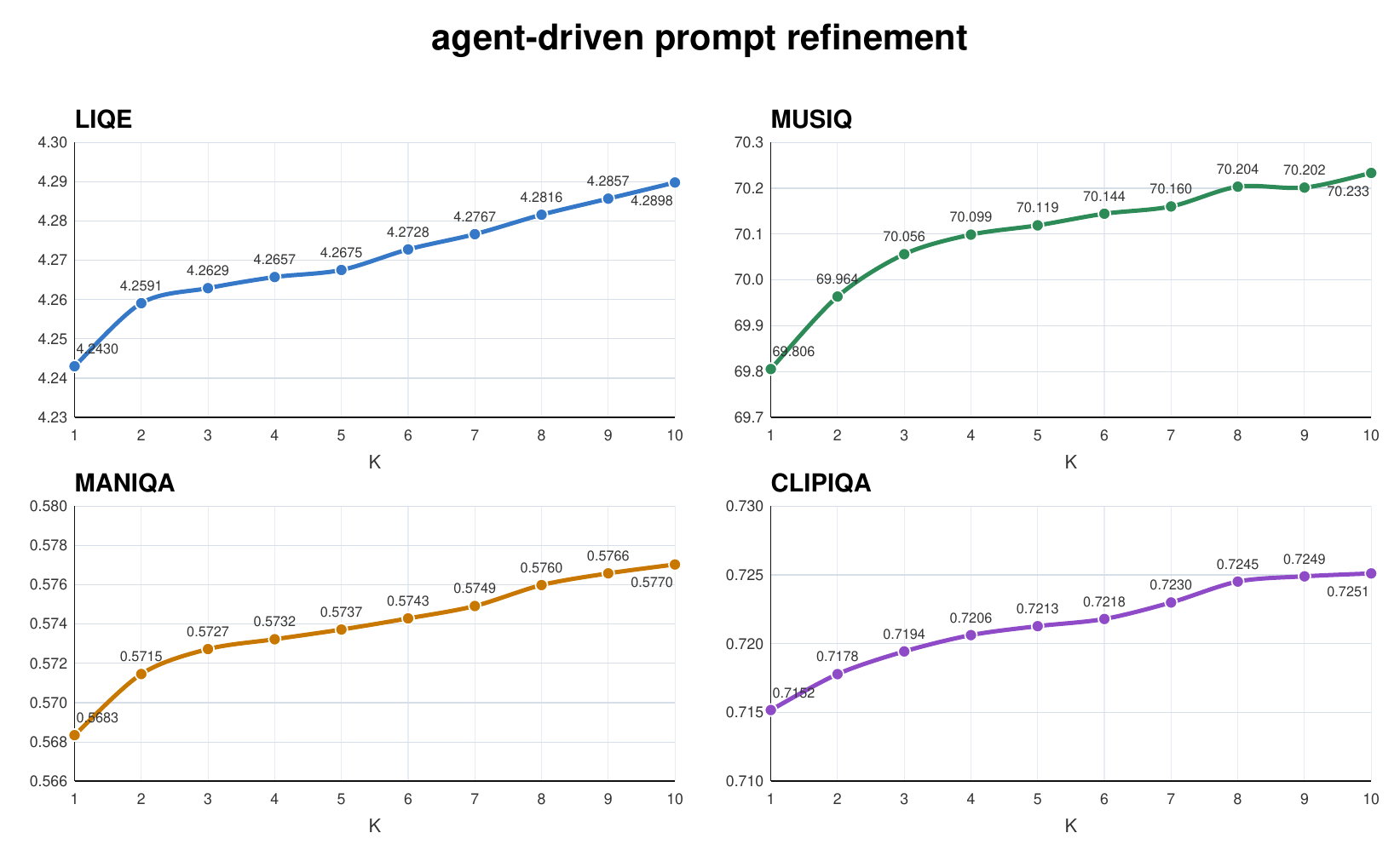}
    \caption{No-reference metric trends of the Best-of-\(K\) selected outputs on RealLR200. \(K=1\) is the adopted static LLaVA initialization; for \(K \ge 2\), one output is selected per image from the first \(K\) generated candidates using the rank-sum rule over LIQE, MUSIQ, MANIQA, and CLIPIQA.}
    \label{fig:appendix_agent_metric_trend}
\end{figure}

Table~\ref{tab:agent_prompt_ablation} reports the practical Best-of-\(K\) variant starting from the adopted static LLaVA initialization. As \(K\) increases, the candidate pool becomes larger and the selected set obtains stronger aggregate no-reference metrics. Figure~\ref{fig:appendix_agent_metric_trend} visualizes this Best-of-\(K\) selection trend. These results support the effectiveness of using inference-time visual feedback to refine semantic prompts for Real-ISR.
The selection rule is fixed before evaluation and applied automatically to every image; no result is manually selected. For \(K=1\), the selected output is the adopted static LLaVA initialization. For \(K \ge 2\), only one output is chosen from the first \(K\) generated candidates for each image, and the final metrics are recomputed on this selected output set. Therefore, the Best-of-\(K\) results should be read together with the raw round-wise results in Table~\ref{tab:agent_refine_single_round_reallr200}.

\begin{figure}[t]
    \centering
    \includegraphics[width=0.85\textwidth]{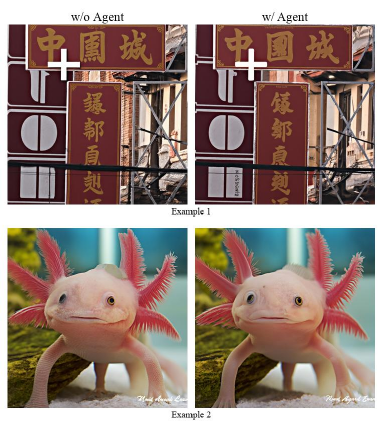}
    \caption{Qualitative examples showing the effect of agent-refined prompting. Each row compares the output without agent refinement and the output with agent refinement. The agent-refined setting produces more coherent semantic details and reduces locally inconsistent structures, supporting the effectiveness of inference-time visual feedback for prompt refinement.}
    \label{fig:appendix_agent_effect}
\end{figure}

Figure~\ref{fig:appendix_agent_effect} provides qualitative examples consistent with the quantitative trend.

\subsection{Qualitative analysis of the global-only variant}

\begin{figure}[t]
    \centering
    \includegraphics[width=0.85\textwidth]{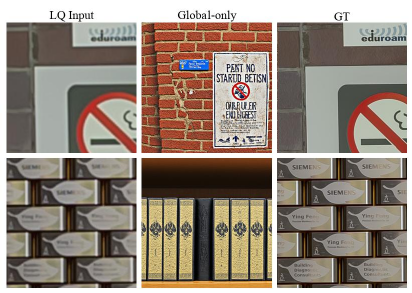}
    \caption{Qualitative examples of the global-only variant. Each row shows the LQ input, the global-only output, and the GT image. Although the global-only variant can produce perceptually plausible images and obtain favorable no-reference scores, it is weakly constrained by the input condition and may synthesize content that is inconsistent with the LQ image and GT.}
    \label{fig:appendix_global_only_fidelity}
\end{figure}

Figure~\ref{fig:appendix_global_only_fidelity} further illustrates why favorable no-reference scores from global context alone should not be interpreted as faithful SR behavior.

\clearpage

\section{Additional Qualitative Results}
\label{app:additional_qualitative_results}

\subsection{Teaser comparison}

\begin{figure*}[t]
    \centering
    \includegraphics[width=\textwidth]{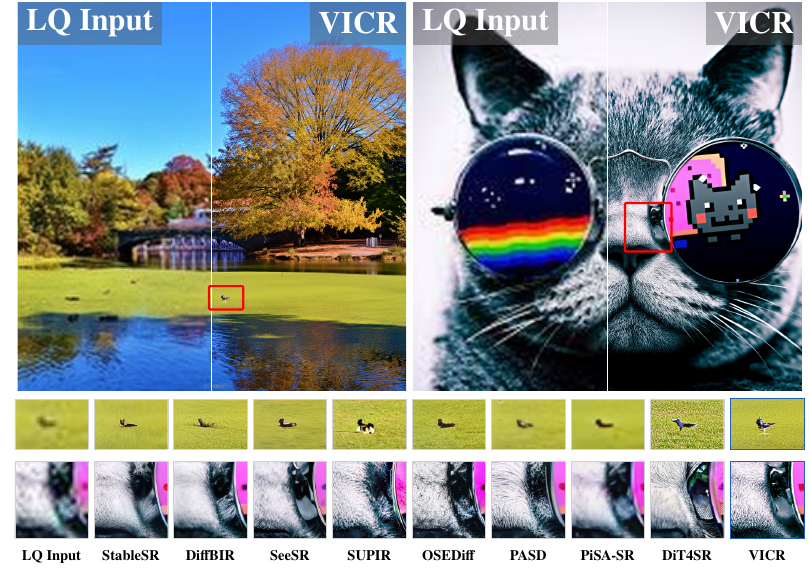}
    \caption{Performance comparison between VICR and other state-of-the-art Real-ISR methods on two challenging real-world LQ images. With a DiT-based generative prior, structured visual conditioning, and semantic prompts refined at inference time by the agent for these examples, VICR restores realistic fine details while remaining faithful to the input, avoiding hallucinated or spatially inconsistent content.}
    \label{fig:teaser}
\end{figure*}

Figure~\ref{fig:teaser} provides an additional teaser-style qualitative comparison.

\subsection{Diverse real-world examples}

\begin{figure*}[t]
    \centering
    \includegraphics[width=1\textwidth]{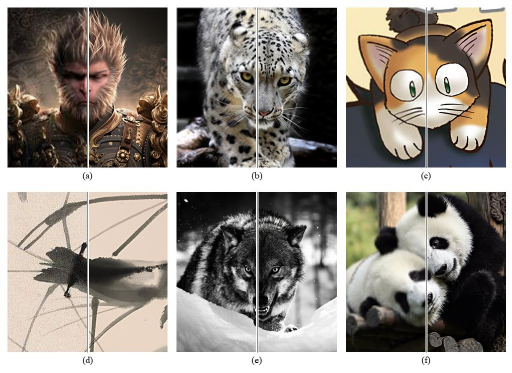}
    \caption{Additional qualitative examples of VICR on diverse low-quality inputs. In each split view, the left half shows the low-quality input upsampled to the output resolution, and the right half shows the VICR output.}
    \label{fig:appendix_vicr_showcase}
\end{figure*}

Figure~\ref{fig:appendix_vicr_showcase} shows additional VICR outputs on diverse low-quality inputs.

\subsection{Additional zoom-in comparisons}

\begin{figure}[t]
    \centering
    \includegraphics[height=0.92\textheight]{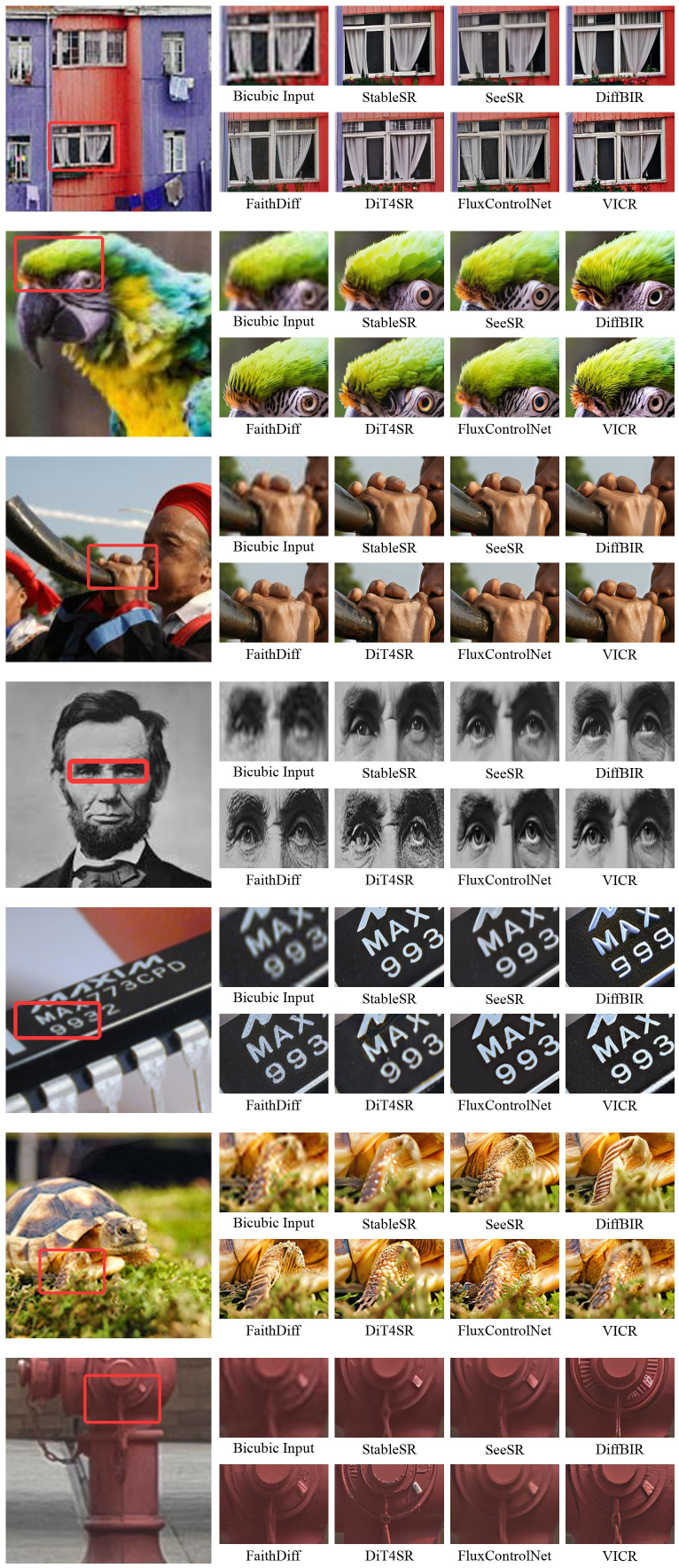}
    \caption{Additional qualitative zoom-in comparisons on challenging low-quality inputs. The examples complement Figure~\ref{fig:qualitative_comparison} by showing more cases where VICR preserves local structures and realistic details under diverse degradations.}
    \label{fig:appendix_more_qualitative}
\end{figure}

Figure~\ref{fig:appendix_more_qualitative} provides additional zoom-in comparisons that complement the main qualitative figure. Across architectural details, animal textures, fingers, faces, printed chip characters, and object parts, VICR more consistently restores recognizable local structures, while several baselines oversmooth details or synthesize sharper but misaligned patterns.

\clearpage

\section{Limitations and Future Work}
\label{app:limitations}

VICR is currently limited by its fixed-resolution visual in-context completion formulation. In our implementation, the model is trained with a diptych composed of two \(512\times512\) images, where the left panel provides the upsampled LQ visible context and the right panel defines the SR target region to be completed. As a result, the model mainly learns context-to-target correspondence under a fixed spatial scale and a fixed left-right layout. For real images with arbitrary resolutions or non-standard aspect ratios, inference therefore requires resizing, cropping, or tiled processing, which may alter the original spatial structure or reduce global context modeling across image regions.

When an input image is divided into multiple tiles, global context can be fragmented across tiles, leading to structural misalignment, texture discontinuities, or low-frequency color inconsistency near tile boundaries. This issue can be more pronounced for large objects, long-range repetitive textures, and semantic regions that extend across several tiles, because each local tile may be insufficient to recover the complete scene relationship independently.

Future work will explore more flexible arbitrary-resolution visual in-context completion mechanisms for Real-ISR, such as multi-scale diptych construction, overlapping tiled inference, and consistency constraints across image tiles. In addition, inspired by spatially focused prompting strategies such as SRSR~\citep{chen2025srsr}, we plan to investigate region-aware textual conditioning in DiT-based Real-ISR\@. This direction may clarify how localized text guidance interacts with regional DiT tokens, how it should be aligned with visual evidence, and whether it can alleviate the context fragmentation caused by fixed-window or tiled inference.

\section{Broader Impacts}
\label{app:broader_impacts}

VICR may benefit applications that require improving the perceptual quality and readability of degraded images, such as image archival, content creation, and visual inspection under low-quality capture conditions. However, as a generative Real-ISR method, VICR may synthesize plausible but not necessarily authentic details in severely degraded regions. Such outputs should therefore not be used as evidence in high-stakes settings such as forensics, medical diagnosis, identity verification, or surveillance without task-specific validation and clear disclosure that the images have been enhanced. These risks motivate cautious deployment and transparent communication of the generative nature and limitations of SR outputs.

\end{document}